\documentclass[journal]{IEEEtran} % Regular and Survey papers
\usepackage{graphicx}
\usepackage{picinpar}
\usepackage{amsmath,mathrsfs,amsthm,bm} % Math packages
\usepackage{stfloats}
\usepackage{url}
\usepackage{flushend}
\usepackage{colortbl}
\usepackage{xcolor,soul,framed} 
\colorlet{shadecolor}{yellow}
\usepackage{multirow}
\usepackage{pifont}
\usepackage{color,soul}
\usepackage{alltt}
\usepackage[hidelinks]{hyperref}
\usepackage{enumerate}
\usepackage{siunitx}
\usepackage{breakurl}
\usepackage{epstopdf}
\usepackage{pbox}
\usepackage{amssymb}
\usepackage{comment}

\usepackage{subfigure}
\usepackage{algorithm}
\usepackage{algpseudocode}

\usepackage{autobreak}
\usepackage[percent]{overpic} % put text on figures
\usepackage{rotating}
\graphicspath{{../figures/}}
\DeclareGraphicsExtensions{.eps,.eps_tex,.pdf,.jpeg,.png}
\definecolor{limegreen}{rgb}{0.2, 0.8, 0.2}
\definecolor{forestgreen}{rgb}{0.13, 0.55, 0.13}
\definecolor{greenhtml}{rgb}{0.0, 0.5, 0.0}
\definecolor{skyblue}{rgb}{0.53, 0.81, 0.92}
\definecolor{lightgray}{rgb}{0.83, 0.83, 0.83}
\definecolor{gray}{rgb}{0.75, 0.75, 0.75}
\definecolor{darkgray}{rgb}{0.66, 0.66, 0.66}
\usepackage{mathtools}
\colorlet{shadecolor}{yellow!255}
\usepackage{filecontents}
\usepackage[noadjust]{cite}
\setstcolor{red}
\renewcommand{\vec}[1]{\mathbf{#1}}

\usepackage{tikz}
\newcommand\copyrighttext{%
  \scriptsize \it This work has been submitted to the IEEE for possible publication. Copyright may be transferred without notice, after which this version may no longer be accessible.}
\newcommand\copyrightnotice{%
\begin{tikzpicture}[remember picture,overlay]
\node[anchor=south,yshift=22pt] at (current page.south) {{\parbox{\dimexpr\textwidth-\fboxsep-\fboxrule\relax}{\copyrighttext}}};
\end{tikzpicture}%
}
\begin{document}
%==================================================
% \title, authors, abstract, and keywords
%==================================================
\title{Redundancy Resolution and Disturbance Rejection via Torque Optimization in Hybrid Cable-Driven Robots}

\author{
	\vskip 1em
    {
    Ronghuai Qi, Amir Khajepour, and William W. Melek, \emph{Senior Member, IEEE}
    }
\thanks{
{This work was supported in part by the Natural Sciences and Engineering Research Council of Canada.

R. Qi, A. Khajepour, and W. W. Melek are with the Department of Mechanical and Mechatronics Engineering, University of Waterloo, Waterloo, ON N2L 3G1, Canada (e-mail: ronghuai.qi@uwaterloo.ca; a.khajepour@uwaterloo.ca; william.melek@uwaterloo.ca).
		}
	}
}

\maketitle
\begin{abstract}
This paper presents redundancy resolution and disturbance rejection via torque optimization in Hybrid Cable-Driven Robots (HCDRs). To begin with, we initiate a redundant HCDR for nonlinear whole-body system modeling and model reduction. Based on the reduced dynamic model, two new methods are proposed to solve the redundancy resolution problem: joint-space torque optimization for actuated joints (TOAJ) and joint-space torque optimization for actuated and unactuated joints (TOAUJ), and they can be extended to other HCDRs. Compared to the existing approaches, this paper provides the first solution (TOAUJ-based method) for HCDRs that can solve the redundancy resolution problem as well as disturbance rejection. Additionally, this paper develops detailed algorithms targeting TOAJ and TOAUJ implementation. A simple yet effective controller is designed for generated data analysis and validation. Case studies are conducted to evaluate the performance of TOAJ and TOAUJ, and the results suggest the effectiveness of the aforementioned approaches.
\end{abstract}


\copyrightnotice

\IEEEpeerreviewmaketitle

%==================================================
% \section{Introduction}
%==================================================
\section{Introduction}
\IEEEPARstart{H}{ybrid} cable-driven robots (HCDRs) are industrial robots that are created by combining cable-driven parallel robots (CDPRs, e.g., in \cite{Mendez2014, H.Jamshidifar2017,H.Jamshidifar2018,Z.Mu2018,M.Chen2018,Otis2009}) and serial robots to overcome their own shortages as well as enhance their advantages (e.g., positioning accuracy and workspace). As one of the important topics in robotics, kinematics is concerned with the motion of the robot's joints in relation to the motion of the robot’s end-effector, including forward kinematics and inverse kinematics. For a CDPR, calculating the mobile platform (end-effector) position by the given cable lengths represents forward kinematics; computing the cable lengths by the given the mobile platform position denotes inverse kinematics. For a serial robot, forward kinematics is used to calculate the position and orientation of the end-effector when the joint angles are provided; inverse kinematics is used to compute the joint angles (the position and orientation of the end-effector are given). While for an HCDR, it includes the above two types of kinematic problems (e.g., the proposed HCDR in \autoref{fig:J4_11DOFHCDRModel}).

In addition, redundancy resolution (kinematic redundancy) is another important topic in kinematics and has existed for years. Strictly speaking, redundant robots do not exist, but given tasks lead to their redundancy~\cite{Chiaverini97}. One may fix this problem regarding the category of robots first: under-actuated robots and fully-actuated robots~\cite{R.Tedrake2019,ARSENAULT20131,LIM20111265} (their strict definitions are provided in~\cite{R.Tedrake2019}). The under-actuation represents the degree of freedom (DOF) of a robot $n$ is more than the number of driven cables/independent joint driven actuators $N$, i.e., $n>N$; the fully-actuation denotes the DOF of a robot $n$ is no more than the number of driven cables/independent joint driven actuators $N$, i.e., $n \le N$. Then, the value of $N-n$ represents the degree of redundancy (DOR). When the redundancy problem exists, there are infinite solutions for inverse kinematics. Generally, approximate methods can be utilized to find numerical solutions, such as Jacobian pseudoinverse \cite{J.Li2017}, Jacobian transpose \cite{Wolovich1984}, cyclic coordinate descent \cite{Wang1991}, damped least squares{~}\cite{Wampler1986,J.Li2017,Nakamura1986}, and quasi-Newton and conjugate gradient{~}\cite{Wang1991,Zhao1994} approaches.

Some studies have tried to find the optimal redundancy resolutions: Barrette \cite{Barrette2005} used the inequality and equality constraints to find a redundancy resolution. Bruckmann \cite{Bruckmann2006} utilized different optimization objective functions to solve redundancy problems. In the past few years, researchers also attempted to use the below optimization methods: for instance, minimum energy \cite{Y.Zhang2007}, minimum norm at acceleration level \cite{Flacco2015}, minimum infinity-norm at velocity level \cite{Tang2001,Zhang2006}, inertia-inverse weighted torque{~}\cite{Y.Zhang2004,Z.Zhang2018}, minimum torque norm{~}\cite{Flacco2015,Y.Zhang2004,Z.Zhang2018}, and minimization infinity-norm torque{~}\cite{Y.Zhang2006,Guo2012,Tang1999}. 

{\color{black}
This paper is motivated by the problem of finding redundancy resolution and rejecting disturbances in joint space for HCDRs. The HCDRs in this paper consist of actuated joints and unactuated joints. When kinematic redundancy occurs, the existing approaches focus on solving the redundancy resolution problem for actuated joints. For example, Flacco{~}\cite{Flacco2015} developed a discrete-time method by minimizing the norm of joint acceleration or joint torque. One of the advantages of the minimum weighted torque norm reported in{~}\cite{Kazerounian1988,S.Ma1994,Flacco2015} was that the motions of robot joints normally stayed bounded. However, unactuated joints (e.g., disturbances applied to the unactuated joints may affect the motion of a robot) are not considered in the related literature. To address this problem, this paper proposes two new methods based on the reduced dynamic model, in which the first method called TOAJ. The basic idea of this method is similar to the existing approaches~ \cite{Kazerounian1988,S.Ma1994,Flacco2015}, but we introduce a new damping gain to stabilize self-motions. Though the TOAJ approach, we can obtain the redundancy resolution acting on actuated joints. The second approach called TOAUJ; using this method, we can not only get the redundancy resolution for actuated joints but also reject disturbances appearing in unactuated joints.} In summary, the main contributions of this paper are highlighted as follows:
\begin{enumerate}
\item Nonlinear whole-body dynamics equations of the HCDR are developed, and model reduction methods are proposed. Based on the reduced dynamic model, two new methods are proposed to solve redundancy resolution: TOAJ and TOAUJ. {\color{black}To the best of the authors’ knowledge, it is the first time that the TOAUJ is proposed.} Compared to the existing methods in~\cite{Kazerounian1988,S.Ma1994,Flacco2015}, TOAUJ can solve the redundancy resolution problem as well as active satisfactory disturbance rejection.
\item Detailed algorithms are provided to implement TOAJ and TOAUJ, and the numerical results suggest the effectiveness of these two methods.
\item {\color{black}A simple yet effective controller is designed for numerical analysis and validation.
}
\end{enumerate}

The rest of this paper is organized as follows: in \autoref{sec:J4_SystemModeling} system modeling is developed, then in \autoref{sec:J4_RedundancyResolution}, the redundancy resolution via joint-space torque optimization is proposed. Control design and numerical results are provided in \autoref{sec:J4_OnlineControl} and \autoref{sec:J4_NumericalResults}, respectively. Finally, conclusions are summarized in \autoref{sec:J4_Conclusions}.

%==================================================
% \section{Problem Definition}, source: thesis 06/15/2019
%==================================================
\section{Problem Definition}
\label{sec:J4_ProblemDefinition}
Consider a general second-order nonlinear system \cite{R.Tedrake2019}:
\begin{align}
    \ddot{\vec{q}} =& f(\vec{q},\dot{\vec{q}},\vec{u},t) = {f_{\cal U}}(\vec{q},\dot{\vec{q}},t) + {f_{\cal A}}(\vec{q},\dot {\vec{q}},t)\vec{u},\nonumber\\
    &{\rm{rank}}({f_{\cal A}}(\vec{q},\dot {\vec{q}},t)) < {n_{\cal A}} + {n_{\cal U}},
    \label{eq:J4_ProblemDef_1}
\end{align}
where $f(\vec{q},\dot {\vec{q}},\vec{u},t)$, ${f_{\cal U}}(\vec{q},\dot {\vec{q}},t)$, and ${f_{\cal A}}(\vec{q},\dot {\vec{q}},t)$ indicate an underactuated system, an unactuated subsystem, and a fully-actuated subsystem, respectively. $\vec{q}=:{[\vec{q}_{\cal A}^T,\vec{q}_{\cal U}^T]^T} \in {\mathbb{R}^{{n_{\cal A}} + {n_{\cal U}}}}$, $\vec{q}_{\cal A} \in {\mathbb{R}^{n_{\cal A}}}$, and $\vec{q}_{\cal U} \in {\mathbb{R}^{n_{\cal U}}}$ represent all joint (generalized) coordinates, actuated joint coordinates, and unactuated joint coordinates, respectively. ${\dot {\vec{q}}} \in {\mathbb{R}^{{n_{\cal A}} + {n_{\cal U}}}}$, $\vec{u}$, and $t$ denote a vector of velocities, a vector of system inputs, and time, respectively.

When a new constraint is introduced into \eqref{eq:J4_ProblemDef_1}, i.e.,
\begin{align}
{\dot {\vec{p}}_e} = {\vec{J}_e}{\dot {\vec{q}}_{\cal A}},\;{n_e} \le {n_{\cal A}},
\label{eq:J4_ProblemDef_2}
\end{align}
with ${\dot {\vec{p}}_e} \in {\mathbb{R}^{n_e}}$ and $\vec{J}_e$ denoting the velocity vector of the end-effector and task Jacobian matrix of a robot, respectively. Eq. \eqref{eq:J4_ProblemDef_2} indicates the redundancy resolution problem (i.e., ${\dot {\vec{p}}_e} \to {\dot {\vec{q}}_{\cal A}}$) of a redundant actuated system by given ${\dot {\vec{p}}_e}$. {\color{black}When the velocity ${\dot {\vec{p}}_e}$ (or position ${\vec{p}_e}$) of the end-effector is given, our objective is to solve \eqref{eq:J4_ProblemDef_1} and \eqref{eq:J4_ProblemDef_2} at the same time. To achieve this goal, we initiate an 11-DOF redundant HCDR (see \autoref{fig:J4_11DOFHCDRModel}) for system modeling, algorithms developing, and case studies.}

%==================================================
% \section{System Modeling}
%==================================================
\section{System Modeling}
\label{sec:J4_SystemModeling}
\subsection{System Configuration}
\label{subsec:J4_HCDRConfig}
\begin{figure}[t]\centering
\vspace{0.1cm}
	\includegraphics[width=87mm]{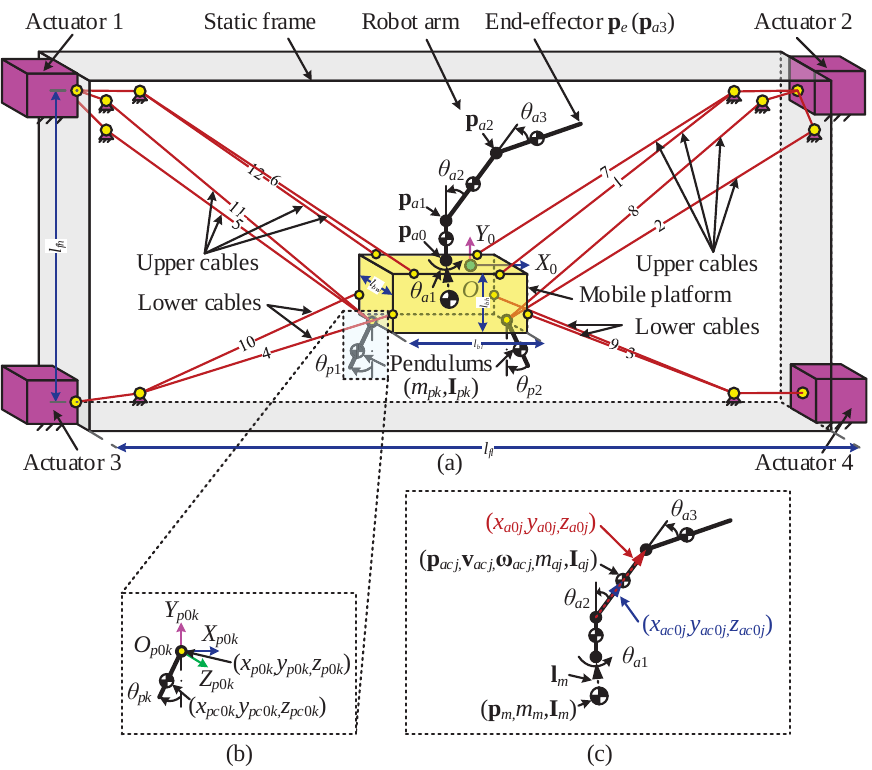}
	\vspace{-0.5cm}
	\caption{The proposed redundant hybrid cable-driven robot (HCDR) consists of a mobile platform, two 1-DOF pendulums, and a 3-DOF robot arm. (a) Overall structure of HCDR. (b) Enlarge view of the pendulum. {\color{black}(c) Additional parameters assignment of the robot arm and moving platform.}}\label{fig:J4_11DOFHCDRModel}
\vspace{-0.7cm}
\end{figure}

The proposed HCDR (\autoref{fig:J4_11DOFHCDRModel}) consists of a mobile platform, two pendulums, and a 3-DOF robot arm. The mobile platform and pendulums are based on the cable-driven platform~\cite{Mendez2014, Rushton2016,Rushton2018,R.Qi2018j2,R.Qi2019j3,R.Qi2019j4}, which consists of a mobile platform, twelve cables, four servo motors, and two 1-DOF pendulums. The actuators are used to drive the cables to move the mobile platform in the $X$-$Y$ plane (i.e., in-plane moving). Twelve cables include four sets of cables: two sets of a four-cable arrangement on the top and two sets of a two-cable arrangement on the bottom, and each set of cables is controlled by one motor. The top actuators and bottom actuators control the upper-cable lengths and lower-cable tensions, respectively. The upper cables also restrict the orientation of the mobile platform, i.e., the kinematic constraints. Meanwhile, some Cartesian coordinate frames are defined as follows: the inertial coordinate frame $O\left\{ {{x_0},{y_0},{z_0}} \right\}$ is located at the center of the static fixture, coordinate frame $\{O_m\}$ is located at the center of mass (COM) of the mobile platform. In addition, two pendulums (mounted on the mobile platform and rotate about their body-fixed $X$-axes) are used to eliminate undesired out-of-plane moving (e.g., vibrations). The robot arm (with its first, second, and third revolute joints rotating about the body-fixed $Y$-, $Z$-, and $Z$-axes, respectively) is mounted on the mobile platform and used for operations such as pick-and-place. {\color{black}The principle of mounting two pendulums on the mobile platform for eliminating out-of-plane vibrations is as below: using four driven cables, the HCDR is in-plane controllable.  However, when the robot arm moves out-of-plane. This movement generates reaction force-moment pairs to the platform resulting in the HCDR is out-of-plane uncontrollable. Two pendulums are introduced to counteract the reaction force and moment pairs. Using two pendulums can lie in the mechanical simplicity and control the remaining DOFs of the platform. More details about design and descriptions are provided in~\cite{Rushton2016,R.Qi2019j3}.}

%\vspace{0.1cm}
\begin{table}[h]
	\renewcommand{\arraystretch}{1.3}
	\vspace{0.1cm}
	\caption{HCDR Parameters}
	\vspace{-0.2cm}
	\centering
	\label{table:J4_11DOFHCDRParameters}
	%\centering
	\resizebox{\columnwidth}{!}{
		\begin{tabular}{c c c c}
			\hline\hline \\[-3mm]
Symbol & Values & Symbol & Values  \\[1.6ex] \hline
$l_{fl}$ & $3.160$ \si{\metre} & $l_{fh}$ & $1.000$ \si{\metre} \\
$l_{bl}$ & $0.365$ \si{\metre}  & $l_{bw}$ & $0.130$ \si[per-mode=symbol]{\metre}  \\
$l_{bh}$ & $0.096$ \si{\metre} & $\vec{l}_{m}=[x_{a0},y_{a0},z_{a0}]^T$  & $[0,0.048,0]^T$ \si{\metre}\\
$m_m$ & \SI{12.200}{\kilogram} & ${\vec{I}}_m$ & $\rm{diag}([0.1021,0.167,0.1251])$ \si{\kilogram\metre\squared}\\
$m_{p1},\;m_{p2}$ & \SI{0.640}{\kilogram} & ${\vec{I}}_{p1},\;{\vec{I}}_{p2}$ & $7.012e{-4}$ \si{\kilogram\metre\squared}\\
$[x_{p01},y_{p01},z_{p01}]^T$ & $[-0.175,-0.100,0]^T$ \si{\metre} & $[x_{p02},y_{p02},z_{p02}]^T$ & $[0.175,-0.100,0]^T$ \si{\metre} \\
$[x_{pc01},y_{pc01},z_{pc01}]^T$ & $[0,-0.050,0]^T$ \si{\metre} & $[x_{pc02},y_{pc02},z_{pc02}]^T$ & $[0,-0.050,0]^T$ \si{\metre}\\
$m_{a1}$ & {$0.300\;\rm{kg}$} & ${\vec{I}}_{a1}$ & {$\rm{diag}([6.76e{-5},6.76e{-5},6.76e{-5}])$} \si{\kilogram\metre\squared}\\
$m_{a2}$, $m_{a3}$ & {$0.300\;\rm{kg}$} & ${\vec{I}}_{a2}$, ${\vec{I}}_{a3}$ & {$\rm{diag}([1.70e{-3},1.70e{-3}, 1.70e{-3}])$} \si{\kilogram\metre\squared}\\
$[x_{a01},y_{a01},z_{a01}]^T$ & {$[0,0.026,0]^T\;\rm{m}$}&
$[x_{ac01},y_{ac01},z_{ac01}]^T$ & {$[0,0.013,0]^T\;\rm{m}$}\\
$[x_{a02},y_{a02},z_{a02}]^T$ & {$[0,0.130,0]^T\;\rm{m}$}&
$[x_{ac02},y_{ac02},z_{ac02}]^T$ & {$[0,0.065,0]^T\;\rm{m}$}\\
$[x_{a03},y_{a03},z_{a03}]^T$ & {$[0,0.130,0]^T\;\rm{m}$}&
$[x_{ac03},y_{ac03},z_{ac03}]^T$ & {$[0,0.065,0]^T\;\rm{m}$}\\
$T_{34\max}$ & $80 \;\rm{N}$&$g$ & $9.810$ \si[per-mode=symbol]{\metre\per\second\squared}\\
${EA}_1$, ${EA}_2$ & $24900 \;\rm{N}$& &\\
		\hline\hline
		\end{tabular}
	}
\vspace{-0.7cm}
\end{table}

\subsection{Nonlinear Whole-Body Dynamics}
\label{subsec:J4_11DOFDynamics}
Consider the HCDR shown in \autoref{fig:J4_11DOFHCDRModel} and define ${{\vec{p}_m}:=[{p_{mx}},{p_{my}},{p_{mz}}]^T} \in {\mathbb{R}^3}$ as the position vector of the coordinate frame $\{O_m\}$ with respect to the coordinate frame $\{O\}$ and ${[{\alpha_m},{\beta_m},{\gamma_m}]^T} \in {\mathbb{R}^3}$ (the orientations of the mobile platform about $X$-, $Y$-, and $Z$-axes, respectively) as the vector of Euler angles. $[{\theta_{p1}},{\theta_{p2}}]^T \in {\mathbb{R}^{2}}$ and $[{\theta_{a1}},{\theta_{a2}},{\theta_{a3}}]^T \in {\mathbb{R}^{3}}$ represent the rotation angles of two 1-DOF pendulums and the 3-DOF robot arm, respectively. Other HCDR parameters used for system modeling are shown in \autoref{table:J4_11DOFHCDRParameters}, where $m_m$ and ${\vec{I}}_m$ represent the mass and moment of inertia of the mobile platform, respectively. $m_{pk}$ and ${\vec{I}}_{pk}$ ($\{\forall \; k \in{\mathbb{N}} :1 \le k \le 2\}$) respectively denote the mass and moment of inertia of the pendulums. The sizes of the static fixture (e.g., $l_{fl}$) and mobile platform (e.g., $l_{bl}$), body-fixed positions (e.g., $[x_{p0k},y_{p0k},z_{p0k}]^T$, $[x_{pc0k},y_{pc0k},z_{pc0k}]^T$, $\{\forall \; k \in{\mathbb{N}} :1 \le k \le 2\}$, $[x_{a0j},y_{a0j},z_{a0j}]^T$, $[x_{ac0j},y_{ac0j},z_{ac0j}]^T$, $\{\forall \; j \in{\mathbb{N}} :1 \le j \le 3\}$), etc. are also given in \autoref{table:J4_11DOFHCDRParameters}. Then, the total kinetic and potential energies are computed as
\begin{align}
{{\cal K}_E} = {}&\frac{1}{2}{m_m}[{{\dot {p}}_{mx}},{{\dot {p}}_{my}},{{\dot {p}}_{mz}}]{[{{\dot {p}}_{mx}},{{\dot {{p}}}_{my}},{{\dot {p}}_{mz}}]^T} + \frac{1}{2}{\bm{\omega}}_m^T{{\vec{I}}_m}{\bm{\omega}_m}\nonumber \\
{}&+ \frac{1}{2}\sum\limits_{k = 1}^2 {\left\{ {{m_{pk}}\vec{v}_{pck}^T{\vec{v}_{pck}} + \bm{\omega}_{pck}^T{{\vec{I}}_{pk}}{\bm{\omega}_{pck}}} \right\}}
 \nonumber \\
{}&+ \frac{1}{2}\sum\limits_{j = 1}^3 {\left\{ {{m_{aj}}\vec{v}_{acj}^T{\vec{v}_{acj}} + \bm{\omega}_{acj}^T{{\vec{I}}_{aj}}{\bm{\omega}_{acj}}} \right\}}
\label{eq:J4_kinetic_energy}
\end{align}
and
\begin{align}
{{\cal V}_E} = {}&{m_m}g{p_{my}} + \sum\limits_{k = 1}^2 {\left\{ {{m_{pk}}g{\vec{p}_{pck}^T}{{[0,1,0]}^T}} \right\}} \nonumber \\
{}&+ \sum\limits_{j = 1}^3 {\left\{ {{m_{aj}}g{\vec{p}_{acj}^T}{{[0,1,0]}^T}} \right\}}\nonumber \\
{}&+\frac{1}{2}{\left( {\vec{L} - {{\vec{L}}_0}} \right)^T}{\vec{K}_c}\left( {\vec{L} - {{\vec{L}}_0}} \right),
\label{eq:J4_potential_energy}
\end{align}
where the term $\frac{1}{2}{\left( {\vec{L} - {{\vec{L}}_0}} \right)^T}{\vec{K}_c}\left( {\vec{L} - {{\vec{L}}_0}} \right)$ denotes the cable elastic potential energy. Details on how other terms in \eqref{eq:J4_kinetic_energy} and \eqref{eq:J4_potential_energy} are calculated are provided in Appendix{~}\ref{sec:J4_appendix_lagrangian_terms}. 

The Lagrangian equation is obtained by substituting \eqref{eq:J4_kinetic_energy} and \eqref{eq:J4_potential_energy} into ${{\cal L}_E} = {{\cal K}_E} - {{\cal V}_E}$. Then, the equations of motion of the HCDR can be derived from ${{\cal L}_E}$ and arranged as
\begin{align}
{{\vec{M}}}\left( {{\vec{q}}} \right){{\ddot {\vec{q}} + {\vec{C}}}}\left( {{{\vec{q}}},{{\dot {\vec{q}}}}} \right){{\dot {\vec{q}} + {\vec{G}}}}\left( {{\vec{q}}} \right) + {\bm{\tau}_d} = 
\begin{bmatrix}
\bm{\tau}_m \\
\bm{\tau}_{p}\\
\bm{\tau}_{a}
\end{bmatrix} 
=
\begin{bmatrix}
\vec{A}_m\vec{T} \\
\bm{\tau}_{p}\\
\bm{\tau}_{a}
\end{bmatrix},
\label{eq:J4_11dof_fwdyn}
\end{align}
where $\vec{q} = [{p_{mx}},{p_{my}},{p_{mz}},{\alpha_m},{\beta_m},{\gamma_m},{\theta_{p1}},{\theta_{p2}},{\theta_{a1}},{\theta_{a2}},{\theta_{a3}}$ $]^T \in {\mathbb{R}^{11}}$, ${{\dot {\vec{q}}}} \in \mathbb{R}{^{11}}$, and ${{\ddot {\vec{q}}}} \in \mathbb{R}{^{11}}$, represent the vectors of generalized coordinates, velocities, and accelerations, respectively. ${{\vec{M}}}\left( {{\vec{q}}} \right) \in \mathbb{R}{^{11 \times 11}}$ denotes the combined inertia matrix, ${{{\vec{C}}}}\left( {{{\vec{q}}},{{\dot {\vec{q}}}}} \right) \in \mathbb{R}{^{11 \times 11}}$ represents the combined Coriolis and centripetal matrix, and ${\vec{G}}(\vec{q}) \in {\mathbb{R}^{11}}$ denote the gravitational vector, respectively. ${\bm{\tau}_d} \in {\mathbb{R}^{11}}$, ${\bm{\tau}_m} \in \mathbb{R}{^6}$, ${\bm{\tau}_{p}} \in \mathbb{R}{^2}$, ${\bm{\tau}_{a}} \in \mathbb{R}{^3}$ are unknown bounded disturbances, forces/torques of the mobile platform, pendulums, and robot arm in the generalized coordinates, respectively. ${\vec{A}_m} \in \mathbb{R}{^{6 \times 12}}$ and ${{\vec{T}}} \in \mathbb{R}{^{12}}$ represent the structure matrix and cable tensions, respectively. ${\vec{A}_m}$ is determined by the position and orientation of the mobile platform \cite{R.Qi2018j2}:
\begin{align}
{\vec{A}_m}=\begin{bmatrix}
{{\hat {\vec{L}}}_1} & \cdots & {{\hat {\vec{L}}}_{12}} \\
{\vec{R}_g^m\begin{bmatrix} r_{1x} \\ r_{1y} \\ r_{1z} \\ \end{bmatrix}} \times {{\hat {\vec{L}}}_1} & \cdots & {\vec{R}_g^m\begin{bmatrix} r_{12x} \\ r_{12y} \\ r_{12z} \\ \end{bmatrix}} \times {{\hat {\vec{L}}}_{12}}\end{bmatrix},
\label{eq:J4_11DOF_Am}
\end{align}
where the rotation matrix $\vec{R}_g^m$ and the $i$th unit cable-length vector ${{\hat {\vec{L}}}}_i$ are computed in Appendix{~}\ref{sec:J4_appendix_lagrangian_terms}.

\subsection{Model Reduction}
\label{subsec:J4_ModelReduction}
Eq.{~}\eqref{eq:J4_11dof_fwdyn} provides the complete nonlinear dynamic model with the 6-DOF mobile platform driven by 12 cables. One can simplify this model by considering the kinematic constraints of the HCDR (shown in \autoref{subsec:J4_HCDRConfig}). The mobile platform is actuated in the $X$-$Y$ plane (in-plane motion) with the nominal angle ${\gamma_m}$ is equal to zero. The twelve cable-driven platform is equivalent to a four cable-driven platform \cite{R.Qi2018j2}, i.e., by given ${L_{01}}$, ${L_{02}}$, ${T_3}$, and ${T_4}$, where ${L_{01}}$ (driven by actuator 1) denotes unstretched cable lengths 5, 6, 11, and 12; ${L_{02}}$ (driven by actuator 2) represents unstretched cable lengths 1, 2, 7, and 8; ${T_3}$ (driven by actuator 3) represents cable tensions 4 and 10; and ${T_4}$ (driven by actuator 4) denotes cable tensions 3 and 9. These cables are labeled in \autoref{fig:J4_11DOFHCDRModel}. The inputs to the whole system are defined as $[{{L_{01}},{L_{02}},{T_3},{T_4},{\tau_{p1}},{\tau_{p2}},{\tau_{a1}},{\tau_{a2}},{\tau_{a3}}]^T} \in {\mathbb{R}^9}$ with ${\bm{\tau}_{p}}={[{\tau_{p1}},{\tau_{p2}}]^T} \in {\mathbb{R}^2}$ denoting input torques of the two pendulums and ${\bm{\tau}_{a}}={[{\tau_{a1}},{\tau_{a2}},{\tau_{a3}}]^T} \in {\mathbb{R}^3}$ representing input torques of the 3-DOF robot arm.

Additionally, in{~}\eqref{eq:J4_11dof_fwdyn}, the redundancy resolution problem (i.e., ${\bm{\tau}_m }={{\vec{A}_m}}\vec{T}$) resulting from multi-cables can be solved as follows: since the equivalent four-cable planar CDPR has one DOR, then ${\vec{A}_m}$ (equals $ {\vec{A}_{m[1,2,6][6,1,4,3]}}$ in \eqref{eq:J4_11dof_fwdyn}) is redefined as a new ${3 \times 4}$ matrix. One can restrict one of lower cable tensions ${T_i}$ (for $i=3,4$) to the maximum allowable value ${T_{34\rm{max}}}$. In this case, the redundancy resolution and optimal cable tensions ${\vec{T}_{{\rm{opt}}}}$ are computed as
\begin{subequations}
\label{eq:J4_4CableOptTension}
\begin{align}
&{\vec{T}_{{\rm{opt}}}} = 
\begin{cases}
{{{[{\Gamma_1},{\Gamma_2},{T_{34\max }},{\Gamma_3}]}^T}\quad {\rm{if}}\;{\Gamma_3} \ge {T_{34\max }}}\\
{{{[{\Gamma_1},{\Gamma_2},{\Gamma_3},{T_{34\max }}]}^T}\quad {\rm{otherwise}}}
\end{cases}, \label{eq:J4_4CableOptTension_a}\\
&{\bm{\tau}_{m[1,2,6]}}={{\vec{A}_m}}
\begin{bmatrix}
{T_{1\rm{opt}}}\\
{T_{2\rm{opt}}}\\
T_3\\
T_4
\end{bmatrix}
={{\vec{A}_m}}
\begin{bmatrix}
\frac{EA_1}{L_{01}}\left(L_1 - L_{01} \right)\\
\frac{EA_2}{L_{02}}\left(L_2 - L_{02} \right)\\
{T_{3\rm{opt}}}+{\delta T_3}\\
{T_{4\rm{opt}}}+{\delta T_4}
\end{bmatrix}, \label{eq:J4_4CableOptTension_b}\\
&{L_{01}} = \frac{{E{A_1}{L_1}}}{{E{A_1} + {T_{{\rm{1opt}}}}}},{L_{02}} = \frac{{E{A_2}{L_2}}}{{E{A_2} + {T_{{\rm{2opt}}}}}},
\label{eq:J4_4CableOptTension_c}
\end{align}
\end{subequations}
where $\bm{\Gamma}  = {({{\bar {\vec{A}}}_{mi}})^{-1}}({[0,({m_m} + {m_{p1}} + {m_{p2}} + {m_{a1}} + {m_{a2}}}$ ${+ {m_{a3}})g,0]^T} - {\vec{A}_{mi}}{T_{34\max }})$, ${{\vec{A}_m}} = [{\vec{A}_{m1}},{\vec{A}_{m2}},{\vec{A}_{m3}},{\vec{A}_{m4}}]$, ${{\bar {\vec{A}}}_{m3}} = [{\vec{A}_{m1}},{\vec{A}_{m2}},{\vec{A}_{m4}}]$, ${{\bar {\vec{A}}}_{m4}} = [{\vec{A}_{m1}},{\vec{A}_{m2}},{\vec{A}_{m3}}],$
${\vec{A}_{mi}} \in \mathbb{R}{^{3}}$,{{~}\rm{and}{~}}${{\bar {\vec{A}}}_{mi}} \in \mathbb{R}{^{3 \times 3}}.$ ${T_{34\max }}$ represents the maximum allowable tensions of lower cables (shown in \autoref{table:J4_11DOFHCDRParameters}). ${EA}_1$ and ${EA}_2$ are the product of the elastic modulus and cross-sectional area of the upper two cables (shown in \autoref{table:J4_11DOFHCDRParameters}). In comparison to~\cite{Rushton2016},~\eqref{eq:J4_4CableOptTension} provides a simpler and more effective method. Moreover, ${\delta T_3},{\delta T_4}$ are control inputs of the CDPR denoting changes of lower cable tensions, and the control inputs of the HCDR are defined as ${\vec{u}}=[{{\delta T_3},{\delta T_4},{\tau_{p1}},{\tau_{p2}},{\tau_{a1}},{\tau_{a2}},{\tau_{a3}}]^T} \in {\mathbb{R}^7}$. {\color{black}Compared to the whole-body dynamics~\eqref{eq:J4_11dof_fwdyn}, the reduced dynamic model in this section (e.g., the system DOF and size of matrix $\vec{A}_m$ are decreased from $11$ DOFs and ${6 \times 12}$ to $7$ DOFs and ${3 \times 4}$, respectively) provides a faster yet effective solution to carry out the algorithm (in~\autoref{sec:J4_RedundancyResolution}) and control strategy (in~\autoref{sec:J4_OnlineControl}).}

%==================================================
% \section{Redundancy Resolution via Torque Optimization}
%==================================================
\section{Redundancy Resolution and Disturbance Rejection via Joint-Space Torque Optimization}
\label{sec:J4_RedundancyResolution}
In practical applications (e.g., pick-and-place), it is interesting to study the Cartesian space trajectory of the robot end-effector, which means only Cartesian position $\vec{p}_e$, velocity $\dot {\vec{p}}_e$, and/or acceleration $\ddot {\vec{p}}_e$ are given. The main goal is to find joint-space commands (e.g., angles and velocities), i.e., the redundancy resolution problem. In this case, two new methods are proposed to solve it: TOAJ and TOAUJ.
\subsection{TOAJ Method}
\label{subsec:J4_MinimumTorqueNorm}
To obtain the optimal actuated-joint torque ${{\bm{\tau}}_{\cal A}} \in {\mathbb{R}^5}$, the cost function is defined as
\begin{subequations} 
\label{eq:J4_LagrangeMultiplier1}
\begin{align}
{\mathop{\rm{min}}\limits_{\bm{\tau}_{\cal A} \in {\mathscr{S}_1}}} &{~}{\Lambda  = \frac{1}{2}\left\| {\vec{M}_{\cal A}^{ - 1}{\bm{\tau}_{\cal A}}} \right\|_2^2}\label{eq:J4_LagrangeMultiplier1_a}\\
{\rm{s. \; t.\;}} &{~} {\mathscr{S}_1}=\Big\{{\mathop{\rm{argmin}}}
\;{{{{\ddot {\vec{p}}}_e} = {\vec{J}_e}{\ddot {\vec{q}}_{\cal A}} + {{\dot {\vec{J}}}_e}{\dot {\vec{q}}_{\cal A}}}}\label{eq:J4_LagrangeMultiplier1_b}\\
&{~}{\rm{s.\;t.\quad}}{\vec{M}_{\cal A}({\vec{q}}_{\cal A})}{\ddot {\vec{q}}_{\cal A}} + {\vec{C}_{\cal A}}({{\vec{q}}_{\cal A}},{\dot {\vec{q}}_{\cal A}}){\dot {\vec{q}}_{\cal A}} + {\vec{G}_{\cal A}}({\vec{q}}_{\cal A}) \nonumber\\
&{~}\quad\quad\;\;\; {+ {{\bm{\tau}_d}_{\cal A}}  = {\bm{\tau}_{\cal A}}} \label{eq:J4_LagrangeMultiplier1_c}\\
&{~}\quad\quad\;\;\; {\bm{\tau}_{\cal A}} = {[{F_{mx}},{F_{my}},{\tau_{a1}},{\tau_{a2}},{\tau_{a3}}]^T},\nonumber\\
&{~}\quad\quad\;\;\; {F_{mx}} = {\bm{\tau}_{m[1]}},{F_{my}} = {\bm{\tau}_{m[2]}}\Big\}, \label{eq:J4_LagrangeMultiplier1_d}
\end{align}
\end{subequations}
where ${\vec{q}}_{\cal A}=[{p_{mx}},{p_{my}},{\theta_{a1}},{\theta_{a2}},{\theta_{a3}}{]^T}, {\dot {\vec{q}}_{\cal A}}$, and $\ddot {\vec{q}}_{\cal A}$ represent the actuated vectors of generalized coordinates, velocities, and accelerations, respectively. ${\vec{J}_e}=\frac{{\partial {\vec{p}_e}}}{{\partial {{\vec{q}}_{\cal A}}}} \in \mathbb{R}{^{3 \times 5}}$, ${{\dot {\vec{J}}}_e} \in \mathbb{R}{^{3 \times 5}}$, ${\vec{p}_e} \in \mathbb{R}{^3}$, and ${{\ddot {\vec{p}}}_e} \in \mathbb{R}{^3}$ are the task Jacobian matrix, the time-derivative of ${\vec{J}_e}$, and the position and acceleration of the end-effector, respectively. The inertia matrix ${{\vec{M}_{\cal A}}}\left( {{{\vec{q}}_{\cal A}}} \right) \in \mathbb{R}{^{5 \times 5}}$, Coriolis and centripetal matrix ${{\vec{C}_{\cal A}}}\left( {{{{\vec{q}}_{\cal A}}},{{\dot {\vec{q}}_{\cal A}}}} \right) \in \mathbb{R}{^{5 \times 5}}$, gravitational vector $\vec{G}_{\cal A}({\vec{q}}_{\cal A}) \in {\mathbb{R}^5}$, and disturbance vector ${{\bm{\tau}_d}_{\cal A}} \in {\mathbb{R}^5}$ are obtained by choosing the corresponding actuated-joint elements in \eqref{eq:J4_11dof_fwdyn} and \eqref{eq:J4_4CableOptTension}. ${F_{mx}}$ and ${F_{my}}$ are two redefined variables (for easier understanding), which represent the forces of the mobile platform in the $X$- and $Y$-directions (generalized coordinates), respectively. Other variables (e.g., ${\bm{\tau}_m}, {\bm{\tau}_{a}}, {p_{mx}}, {p_{my}}, {\theta_{a1}}, {\theta_{a2}}, {\theta_{a3}}$) are also defined in \eqref{eq:J4_11dof_fwdyn}.

The Lagrangian function of \eqref{eq:J4_LagrangeMultiplier1} is conducted as
\begin{align}
{\tilde \Lambda} = &{~} {\frac{1}{2}\left\| {\vec{M}_{\cal A}^{ - 1}{\bm{\tau}_{\cal A}}} \right\|_2^2}+ {{\bm{{\bm{\lambda}}}} ^T}({{\ddot {\vec{p}}}_e} - {{\dot {\vec{J}}}_e}{\ddot {\vec{q}}_{\cal A}} - {\vec{J}_e}{\dot {\vec{q}}_{\cal A}})\nonumber \\
= &{~}\frac{1}{2}{({\vec{M}_{\cal A}}{\ddot {\vec{q}}_{\cal A}} + {\vec{C}_{\cal A}}{\dot {\vec{q}}_{\cal A}} + {\vec{G}_{\cal A}} +{{\bm{\tau}_d}_{\cal A}} )^T}{\vec{M}_{\cal A}^{ - 2}}\nonumber\\
&{~}({\vec{M}_{\cal A}}{\ddot {\vec{q}}_{\cal A}} + {\vec{C}_{\cal A}}{\dot {\vec{q}}_{\cal A}}
 + {\vec{G}_{\cal A}} +{{\bm{\tau}_d}_{\cal A}} ) \nonumber \\
&{~}+ {{\bm{{\bm{\lambda}}}} ^T}({{\ddot {\vec{p}}}_e} - {\vec{J}_e}{\ddot {\vec{q}}_{\cal A}} - {{\dot {\vec{J}}}_e}{\dot {\vec{q}}_{\cal A}})\nonumber \\
 =&{~} \frac{1}{2}{{\ddot {\vec{q}}_{\cal A}}^T}{\ddot {\vec{q}}_{\cal A}} + {(\vec{C}_{\cal A}{\dot {\vec{q}}_{\cal A}} + \vec{G}_{\cal A} +{{\bm{\tau}_d}_{\cal A}} )^T}{\vec{M}_{\cal A}^{ - 1}}{\ddot {\vec{q}}_{\cal A}}\nonumber \\
 &{~}+ \frac{1}{2}{({\vec{C}_{\cal A}}{\dot {\vec{q}}_{\cal A}} + {\vec{G}_{\cal A}} +{{\bm{\tau}_d}_{\cal A}} )^T}{\vec{M}_{\cal A}^{ - 2}}({\vec{C}_{\cal A}}{\dot {\vec{q}}_{\cal A}} \nonumber \\
 &{~}+{\vec{G}_{\cal A}} +{{\bm{\tau}_d}_{\cal A}} ) + {{\bm{{\bm{\lambda}}}} ^T}({{\ddot {\vec{p}}}_e} - {\vec{J}_e}{\ddot {\vec{q}}_{\cal A}} - {{\dot {\vec{J}}}_e}{\dot {\vec{q}}_{\cal A}}),
\label{eq:J4_LagrangeMultiplier2}
\end{align}
where ${\bm{{\bm{\lambda}}}}$ represents the Lagrange multiplier. Then, the necessary and sufficient conditions \cite{Luenberger2008} for a minimum  of \eqref{eq:J4_LagrangeMultiplier2} can be computed as
\begin{align}
\begin{cases} 
    \frac{{\partial \tilde \Lambda }}{{\partial {\ddot {\vec{q}}_{\cal A}}}}  &= {\ddot {\vec{q}}_{\cal A}} + {\vec{M}_{\cal A}^{ - 1}}({\vec{C}_{\cal A}}{\dot {\vec{q}}_{\cal A}} + {\vec{G}_{\cal A}} +{{\bm{\tau}_d}_{\cal A}} ) \\
    &{\quad} - \vec{J}_e^T{\bm{{\bm{\lambda}}}}  = \mathbf{0}\\
    \frac{{{\partial ^2}\tilde \Lambda }}{{{\partial ^2}{\ddot {\vec{q}}_{\cal A}}}}  &= \vec{I} > \mathbf{0}\\
    \frac{{\partial \tilde \Lambda }}{{\partial {\bm{\lambda}} }} &= {{\ddot {\vec{p}}}_e} - {\vec{J}_e}{\ddot {\vec{q}}_{\cal A}} - {{\dot {\vec{J}}}_e}{\dot {\vec{q}}_{\cal A}} = \mathbf{0}
\end{cases}.
\label{eq:J4_LagrangeMultiplier3}
\end{align}

By arranging \eqref{eq:J4_LagrangeMultiplier3}, the solution is described as
\begin{align}
{\ddot {\vec{q}}_{\cal A}} = &{~}\vec{J}_e^T{({\vec{J}_e}\vec{J}_e^T)^{ - 1}}({{\ddot {\vec{p}}}_e} - {{\dot {\vec{J}}}_e}{\dot {\vec{q}}_{\cal A}}) - (\vec{I} - \vec{J}_e^T{({\vec{J}_e}\vec{J}_e^T)^{ - 1}}{\vec{J}_e})\nonumber\\
&{~}{\vec{M}_{\cal A}^{ - 1}}({\vec{C}_{\cal A}}{\dot {\vec{q}}_{\cal A}} + {\vec{G}_{\cal A}} +{{\bm{\tau}_d}_{\cal A}} ).
\label{eq:J4_LagrangeMultiplier4}
\end{align}

Eq. \eqref{eq:J4_LagrangeMultiplier4} shows the solution at acceleration level (by given ${{\ddot {\vec{p}}}_e}$). Alternately, it is easy to convert \eqref{eq:J4_LagrangeMultiplier4} into the discrete-time expression by using ${\ddot {\vec{p}}_e}(k) = \frac{{{{\dot {\vec{p}}}_e}(k) - {{\dot {\vec{p}}}_e}(k - 1)}}{{{T_s}}}$, ${{\ddot {\vec{q}}_{\cal A}}}(k) = \frac{{{{{\dot {\vec{q}}_{\cal A}}}}(k) - {{{\dot {\vec{q}}_{\cal A}}}}(k - 1)}}{{{T_s}}}$, and ${\dot {\vec{J}}_e}(k) = \frac{{{\vec{J}_e}(k) - {\vec{J}_e}(k - 1)}}{{{T_s}}}$, in which $T_s$ denotes the sampling time. Then, the recursive formula can be described as % (at the discrete-time velocity level \cite{Flacco2015})
\begin{align}
{\dot {\vec{q}}_{\cal A}}(k) = &{~}\vec{J}_e^T(k){[{\vec{J}_e}(k)\vec{J}_e^T(k)]^{ - 1}}{{\dot {\vec{p}}}_e}(k) + \Big\{\vec{I} - \vec{J}_e^T(k)[{\vec{J}_e}(k)\nonumber\\
&{~}\vec{J}_e^T(k)]^{-1}{\vec{J}_e}(k)\Big\}\Big\{\big\{\vec{I}-{T_s}{[{\vec{M}_{\cal A}}({{\vec{q}}_{\cal A}}(k))]^{-1}}\nonumber\\
&{~}[{\vec{C}_{\cal A}}({{\vec{q}}_{\cal A}}(k),{\dot {\vec{q}}_{\cal A}}(k - 1))]\big\}{\dot {\vec{q}}_{\cal A}}(k - 1)- {T_s}\nonumber\\
&{~}{[{\vec{M}_{\cal A}}({{\vec{q}}_{\cal A}}(k))]^{-1}}
[{\vec{G}_{\cal A}}({{\vec{q}}_{\cal A}}(k))+{{\bm{\tau}_d}_{\cal A}} (k)]\Big\},
\label{eq:J4_LagrangeMultiplier5}
\end{align}
where ${\dot {\vec{p}}_e}$ is the input velocity of the end-effector. \eqref{eq:J4_LagrangeMultiplier4} and \eqref{eq:J4_LagrangeMultiplier5} can be improved by introducing a damping gain ${\vec{K}_{dp{\cal A}} \; (\vec{K}_{dp{\cal A}} \ge \mathbf{0})}$ to stabilize self-motions:
\begin{align}
{\ddot {\vec{q}}_{\cal A}} = &{~}\vec{J}_e^T{({\vec{J}_e}\vec{J}_e^T)^{ - 1}}({{\ddot {\vec{p}}}_e} - {{\dot {\vec{J}}}_e}{\dot {\vec{q}}_{\cal A}}) - (\vec{I} - \vec{J}_e^T{({\vec{J}_e}\vec{J}_e^T)^{ - 1}}{\vec{J}_e})\nonumber\\
&{~}{\vec{M}_{\cal A}^{ - 1}}({\vec{C}_{\cal A}}{\dot {\vec{q}}_{\cal A}} + {\vec{G}_{\cal A}} +{{\bm{\tau}_d}_{\cal A}} -{\vec{K}_{dp{\cal A}} }{{\dot {\vec{q}}_{\cal A}}})
\label{eq:J4_LagrangeMultiplier6}
\end{align}
and
\begin{align}
{\dot {\vec{q}}_{\cal A}}(k) = &{~}\vec{J}_e^T(k){[{\vec{J}_e}(k)\vec{J}_e^T(k)]^{ - 1}}{{\dot {\vec{p}}}_e}(k) + \Big\{\vec{I} - \vec{J}_e^T(k)[{\vec{J}_e}(k)\nonumber\\
&{~}\vec{J}_e^T(k)]^{-1}{\vec{J}_e}(k)\Big\}\Big\{\big\{\vec{I}-{T_s}{[{\vec{M}_{\cal A}}({{\vec{q}}_{\cal A}}(k))]^{-1}}\nonumber\\
&{~}[{\vec{C}_{\cal A}}({{\vec{q}}_{\cal A}}(k),{\dot {\vec{q}}_{\cal A}}(k - 1))] +{T_s}{\vec{K}_{dp{\cal A}} }\big\}\dot {{\vec{q}}_{\cal A}}(k - 1)\nonumber\\
&{~}-{T_s}{[{\vec{M}_{\cal A}}({{\vec{q}}_{\cal A}}(k))]^{-1}}[{\vec{G}_{\cal A}}({{\vec{q}}_{\cal A}}(k))+{{\bm{\tau}_d}_{\cal A}} (k)]\Big\}.
\label{eq:J4_LagrangeMultiplier7}
\end{align}

Eqs. \eqref{eq:J4_LagrangeMultiplier6} and \eqref{eq:J4_LagrangeMultiplier7} can be extended to other redundant robots (i.e., replacing by the corresponding parameters $\vec{J}_e$, ${\vec{M}_{\cal A}}$, etc.). Additionally, for the {HCDR-4} shown in \autoref{fig:J4_11DOFHCDRModel}, two types of motion (${\vec{p}_e}=:{[{p_{ex}},{p_{ey}},{p_{ez}}]^T}$) are available to the end-effector, i.e., 
\begin{align}
\begin{cases}
{p_{ez}} = 0\quad \text{in-plane motion}\\
{p_{ez}} \ne 0\quad \text{out-of-plane motion}
\end{cases},
\label{eq:J4_InOutPlaneMotion}
\end{align}
where ${p_{ex}}$, ${p_{ey}}$, and ${p_{ez}}$ represent the positions in the $X$-, $Y$-, and $Z$-directions (with respect to frame $\{O\}$). When ${p_{ez}} = 0$, the redundancy resolution problem can be solved by using \eqref{eq:J4_LagrangeMultiplier6} or \eqref{eq:J4_LagrangeMultiplier7}. However, when ${p_{ez}} \ne 0$, the constraints of pendulums are needed to associate with \eqref{eq:J4_LagrangeMultiplier1}, i.e., to balance reaction forces/moments generated by the movement of the robot arm. %In this case, the nominal position and angles ${p_{mz}},{\alpha_m},{\beta_m},{\gamma_m}$ are assumed to be zero for simplification.
In this case, an equilibrium condition is considered via the following method: computing the nominal angles of pendulums $({\theta_{p1}},{\theta_{p2}})$ using the obtained 
$({\theta_{a1}},{\theta_{a2}},{\theta_{a3}})$ in the previous step, and the problem is described as
\begin{subequations}
\label{eq:J4_PenduNominalAngle}
\begin{align}
\begin{bmatrix}{\theta_{p1}}\\{\theta_{p2}} \end{bmatrix}=  {\mathop{\rm{argmin}}\limits_{{\theta_{a1}},{\theta_{a2}},{\theta_{a3}}}} &{\quad}{\left[ {\sum {{{\cal M}_y}} ,\sum {{{\cal M}_x}} } \right]{\left[ {\sum {{{\cal M}_y}} ,\sum {{{\cal M}_x}} } \right]^T}} \label{eq:J4_PenduNominalAngle_a}\\
{\rm{s.\;t.}} &{\quad}{\sum {{{\cal M}_x}}}  = {{\cal M}_{ax}} - {{\cal M}_{p1x}} - {{\cal M}_{p2x}}\label{eq:J4_PenduNominalAngle_b}\\
&{\quad} {\sum {{{\cal M}_y}}}  = {{\cal M}_{ay}} - {{\cal M}_{p1y}} - {{\cal M}_{p2y}},\label{eq:J4_PenduNominalAngle_c}
\end{align}
\end{subequations}
where ${{\cal M}_{ax}},\;{{\cal M}_{ay}},\;{{\cal M}_{p1x}},\;{{\cal M}_{p1y}},\;{{\cal M}_{p2x}},{{~}\rm{and}{~}} {{\cal M}_{p2y}}$ denote reaction moments of the robot arm and two pendulums to the mobile platform about its $X$- and $Y$-axes, respectively. These terms can be computed as
\begin{align}
\label{eq:J4_PenduNominalAngle_term1}
\begin{cases}
{{{\cal M}_{ax}} = \vec{M}_a^T{{[1,0,0]}^T},{{\cal M}_{ay}} = \vec{M}_a^T{{[0,1,0]}^T},{\vec{M}_a} = }\\
{\sum\limits_{\ell  = 2}^3 {\left( {({\vec{p}_{a\ell }} - {\vec{p}_{a(\ell  - 1)}})} \right. \times \sum\limits_{j = \ell }^3 {\left. {{\vec{f}_{a(j + 1)}}} \right) + ({\vec{p}_{a1}} - [{p_{mx}},{p_{my}},} } }\\
{{p_{mz}}{]^T}) \times \sum\limits_{j = 1}^3 {{\vec{f}_{a(j + 1)}}}  + \sum\limits_{j = 2}^3 {\left( {({\vec{p}_{acj}} - {\vec{p}_{a(j - 1)}}) \times {\vec{f}_{aj}}} \right)}}\\
 + ({\vec{p}_{ac1}} - {[{p_{mx}},{p_{my}},{p_{mz}}]^T}) \times {\vec{f}_{a1}}\\
 + \sum\limits_{j = 1}^3 {{\vec{R}_{aj}}\left( {{{\vec{I}}_{aj}}{{\dot {\bm{\omega}} }_{acj}} + {\bm{\omega}_{acj}} \times ({{\vec{I}}_{aj}}{\bm{\omega}_{acj}})} \right)}
\end{cases}
\end{align}
and
\begin{align}
\label{eq:J4_PenduNominalAngle_term2}
\begin{cases}
{{\cal M}_{pkx}} = {\cal {\vec{M}}}_{pk}^T{[1,0,0]^T},
{{\cal M}_{pky}} = {\cal {\vec{M}}}_{pk}^T{[0,1,0]^T},{{\cal {\vec{M}}}_{pk}}= \\
({\vec{p}_{pck}} - {[{p_{mx}},{p_{my}},{p_{mz}}]^T}) \times \left( {{m_{pk}}({{\dot {\vec{v}}}_{pck}} + {{[0,g,0]}^T})} \right)\\
+ {\vec{R}_x}({\theta_{pk}})\left( {{{\vec{I}}_{pk}}{{\dot {\bm{\omega}} }_{pck}} + {\bm{\omega}_{pck}} \times ({{\vec{I}}_{pk}}{\bm{\omega}_{pck}})} \right),\;{\rm{for}}\;k = 1,2
\end{cases},
\end{align}
where ${\vec{f}_{aj}} = {m_{aj}}({\dot {\vec{v}}_{acj}} + {[0,g,0]^T}),{\vec{R}_{a1}} = {\vec{R}_y}({\theta_{a1}}),{\vec{R}_{a2}} = {\vec{R}_y}({\theta_{a1}}){\vec{R}_z}({\theta_{a2}}),{\vec{R}_{a3}} = {\vec{R}_y}({\theta_{a1}})$ ${\vec{R}_z}({\theta_{a2}}){\vec{R}_z}({\theta_{a3}}),\;{\rm{and}}\;{\vec{f}_{a4}} = {\bf{0}}$. ${[{p_{mx}},{p_{my}},{p_{mz}}]^T}$ is the position vector of the COM of the mobile platform. For $j = 1,2,3$ and $k = 1,2$, the position, linear velocity, and angular velocity vectors ${\vec{p}_{aj}},{\vec{p}_{acj}},{\vec{p}_{pck}},{\vec{v}_{acj}},{\vec{v}_{pck}},{\bm{\omega}_{acj}},\;{\rm{and}}\;{\bm{\omega}_{pck}}$ are obtained using the equations shown in Appendix{~}\ref{sec:J4_appendix_lagrangian_terms}. The vectors of linear acceleration and angular acceleration ${\dot {\vec{v}}_{acj}},{\dot {\vec{v}}_{pck}},{\dot {\bm{\omega}}_{acj}},\;{\rm{and}}\;{\dot {\bm{\omega}}_{pck}}$ are time-derivatives of ${\vec{v}_{acj}},{\vec{v}_{pck}},{\bm{\omega}_{acj}},\;{\rm{and}}\;{\bm{\omega}_{pck}}$, respectively. Other parameters such as ${m_{aj}},{{\vec{I}}_{aj}},{m_{pk}},{{\vec{I}}_{pk}},\;{\rm{and}}\;g$ are provided in{~}\autoref{table:J4_11DOFHCDRParameters}. Eq.{~}\eqref{eq:J4_PenduNominalAngle} is a nonlinear optimization problem and can be solved using nonlinear solvers (e.g., MATLAB function \textit{fmincon} which is used for case studies in~\autoref{sec:J4_NumericalResults}).

%--------------------------------------------------
% 
%--------------------------------------------------
\subsection{TOAUJ Method}\label{subsec:J4_AugmentedMinTorque}
The cost function \eqref{eq:J4_LagrangeMultiplier1_a} is used to solve the redundancy resolution problem by minimizing actuated joint torques. However, when unactuated joints exist, minimum of actuated joint torques may not be guaranteed due to the coupled actuated and unactuated joints, e.g., disturbances resulting from unactuated joints. In this case, a new cost function is proposed to address this problem:
\begin{subequations} 
\label{eq:J4_AugMinTorqueCostFun}
\begin{align}
{\mathop{\rm{min}}\limits_{{\bm{\tau}_{\cal A}},{\bm{\tau}_{\cal U}} \in {\mathscr{S}_2}}} &{~}{\Lambda  = \frac{1}{2}\left\| {\vec{M}_{{\cal A}{\cal U}}^{ - 1}{{[\bm{\tau}_{\cal A}^T,\bm{\tau}_{\cal U}^T]}^T}} \right\|_2^2} \label{eq:J4_PenduNominalAngle_a}\\
{\rm{s. \; t.\;}} &{~} {\mathscr{S}_2}=\Big\{{\mathop{\rm{argmin}}}
\;{{{{\ddot {\vec{p}}}_e} = {\vec{J}_e}{\ddot {\vec{q}}_{\cal A}} + {{\dot {\vec{J}}}_e}{\dot {\vec{q}}_{\cal A}}}}\label{eq:J4_PenduNominalAngle_a2b}\\
&{~}{\rm{s.\;t.\quad}}{\bm{\tau}_{\cal A}} = {[{F_{mx}},{F_{my}},{\tau_{a1}},{\tau_{a2}},{\tau_{a3}}]^T}
\label{eq:J4_AugMinTorqueCostFun_b}\\
&{~}\quad\quad\;\;\; {\bm{\tau}_{\cal U}} = {[{F_{mz}},{M_{mx}},{M_{my}}]^T}\label{eq:J4_AugMinTorqueCostFun_c}\\
&{~}\quad\quad\;\;\; {F_{mx}} = {{\bm{\tau}}_{m[1]}},{F_{my}} = {{\bm{\tau}}_{m[2]}},{F_{mz}} = {{\bm{\tau}}_{m[3]}},\nonumber\\
&{~}\quad\quad\;\;\; {M_{mx}} = {{\bm{\tau}}_{m[4]}},{M_{my}} = {{\bm{\tau}}_{m[5]}} \Big\}, \label{eq:J4_AugMinTorqueCostFun_d}
\end{align}
\end{subequations}
where ${\bm{\tau}_{\cal A}}$, ${\bm{\tau}_{\cal U}}$, and ${{\vec{M}}_{{\cal A}{\cal U}}}$ denote the actuated torque vector, unactuated torque vector, and combined inertia matrix, respectively. The forces $({F_{mx}},{F_{my}},{F_{mz}})$ and torques $({M_{mx}},{M_{my}},{\tau_{a1}},{\tau_{a2}},{\tau_{a3}})$ in \eqref{eq:J4_AugMinTorqueCostFun_d} of the mobile platform and robot arm are obtained using{~}\eqref{eq:J4_LagrangeMultiplier1_c} and{~}\eqref{eq:J4_LagrangeMultiplier1_d}. Then, the new mapping from actuated and unactuated joints to the end-effector (velocity vector ${{\dot {\vec{p}}}_e}$ and acceleration vector ${{\ddot {\vec{p}}}_e}$) is computed as
\begin{align}
\begin{cases}
{{\dot {\vec{p}}}_e} = [{\vec{J}_e},{\bf{0}}]{[\dot {\vec{q}}_{\cal A}^T,\dot {\vec{q}}_{\cal U}^T]^T}\\
{{\ddot {\vec{p}}}_e} = [{\vec{J}_e},{\bf{0}}]{[\ddot {\vec{q}}_{\cal A}^T,\ddot {\vec{q}}_{\cal U}^T]^T} + [{{\dot {\vec{J}}}_e},{\bf{0}}]{[\dot {\vec{q}}_{\cal A}^T,\dot {\vec{q}}_{\cal U}^T]^T}
\end{cases},
\label{eq:J4_AugJacbMapping}
\end{align}
where ${\vec{q}}_{\cal A}=[{p_{mx}},{p_{my}},{\theta_{a1}},{\theta_{a2}},{\theta_{a3}}{]^T}$, $\dot {\vec{q}}_{\cal A}$, $\ddot {\vec{q}}_{\cal A}$, ${\vec{q}}_{\cal U}=[{p_{mz}},{\alpha_m},{\beta_m}{]^T}$, $\dot {\vec{q}}_{\cal U}$, and $\ddot {\vec{q}}_{\cal U}$ represent the actuated and unactuated vectors of generalized coordinates, velocities, and accelerations, respectively.

One can also conduct the Lagrangian function (in the form of \eqref{eq:J4_LagrangeMultiplier2}) to solve \eqref{eq:J4_AugMinTorqueCostFun} and the discrete solution is described as
\begin{align}
\begin{bmatrix}{{{\dot {\vec{q}}}_{\cal A}}(k)}\\{{{\dot {\vec{q}}}_{\cal U}}(k)}\end{bmatrix}=&
\begin{bmatrix} {\vec{J}_e^ + {{\dot {\vec{p}}}_e}(k)}\\{\bf{0}}\end{bmatrix}+
\begin{bmatrix} {{{\tilde {\vec{J}}}_e}({{\vec{I}}_{5 \times 5}} + {T_s}{\vec{K}_{dp{\cal A}}})} \qquad{\bf{0}}\\{\bf{0}} \qquad\quad\;\;\;{{{\vec{I}}_{3 \times 3}} + {T_s}{\vec{K}_{dp{\cal U}}}}
\end{bmatrix}\nonumber\\
&\begin{bmatrix} {{{\dot {\vec{q}}}_{\cal A}}(k - 1)}\\{{{\dot {\vec{q}}}_{\cal U}}(k - 1)}\end{bmatrix}
-\begin{bmatrix}{{{\tilde {\vec{J}}}_e}}&{\bf{0}}\\
{\bf{0}}&{{{\vec{I}}_{3 \times 3}}}\end{bmatrix}{T_s}{\vec{M}}_{{\cal A}{\cal U}}^{ - 1}\begin{bmatrix}{{\vec{F}_{\cal A}}}\\{{\vec{F}_{\cal U}}}\end{bmatrix},
\label{eq:J4_AugMinTorqueSol1}
\end{align}
with
\begin{subequations}
\label{eq:J4_AugMinTorqueSol1_1}
\begin{align}
{{\tilde {\vec{J}}}_e}:= &{~}{{{\vec{I}}_{5 \times 5}} - {\vec{J}_e^+}(k) {\vec{J}_e}(k)},\label{eq:J4_AugMinTorqueSol1_1_a}\\
{\vec{F}_{\cal A}}:= &{~}{{[{\vec{C}}(\vec{q}(k),\dot {\vec{q}}(k - 1))]_{[1,2,9:11]}}{{\dot {\vec{q}}}_{\cal A}}(k - 1)}\nonumber\\
&{~}{+ {[{\vec{G}}(\vec{q}(k)) + {\bm{\tau}_d}(k)]_{[1,2,9:11]}}},\label{eq:J4_AugMinTorqueSol1_1_b}\\
{\vec{F}_{\cal U}}:= &{~}{{[{\vec{C}}(\vec{q}(k),\dot {\vec{q}}(k - 1))]_{[3:5]}}{{\dot {\vec{q}}}_{\cal U}}(k - 1)}\nonumber\\
&{~}{+ {[{\vec{G}}(\vec{q}(k)) + {\bm{\tau}_d}(k)]_{[3:5]}}},\label{eq:J4_AugMinTorqueSol1_1_c}
\end{align}
\end{subequations}
where $\vec{J}_e^+$ is the pseudoinverse of matrix $\vec{J}_e$. Finally, \eqref{eq:J4_AugMinTorqueSol1} can be rearranged as
\begin{subequations} 
\label{eq:J4_AugMinTorqueSol2}
\begin{align}
{{\dot {\vec{q}}}_{\cal A}}(k) = &{~}{\vec{J}_e^ + {{\dot {\vec{p}}}_e}(k) + {{\tilde {\vec{J}}}_e}(\vec{I} + {T_s}{\vec{K}_{dp{\cal A}}}){{\dot {\vec{q}}}_{\cal A}}(k - 1)}\nonumber\\
&{~}{- {T_s}({{\vec{I}}_{5 \times 5}} - \vec{J}_e^ + {\vec{J}_e}){\bm{\Xi}_{\cal A}}}, \label{eq:J4_AugMinTorqueSol2_a}\\
{{\dot {\vec{q}}}_{\cal U}}(k) = &{~}({{\vec{I}}_{3 \times 3}} + {T_s}{\vec{K}_{dp{\cal U}}}){{\dot {\vec{q}}}_{\cal U}}(k - 1) - {T_s}{\bm{\Xi}_{\cal U}}, \label{eq:J4_AugMinTorqueSol2_b}
\end{align}
\end{subequations}
with
\begin{subequations}
\label{eq:J4_AugMinTorqueSol2_1}
\begin{align}
\begin{bmatrix}
{\vec{M}_{{\cal A}{\cal U}}^{{\cal A}{\cal A}}}&{\vec{M}_{{\cal A}{\cal U}}^{{\cal A}{\cal U}}}\\
{\vec{M}_{{\cal A}{\cal U}}^{{\cal U}{\cal A}}}&{\vec{M}_{{\cal A}{\cal U}}^{{\cal U}{\cal U}}}
\end{bmatrix}:=&{~}
{\left[ {\begin{matrix}
{{\vec{M}_{{\cal A}{\cal U}[1,2,6:8]  [1,2,6:8]}}(k)},\\
{{\vec{M}_{{\cal A}{\cal U}[3:5]  [1,2,6:8]}}(k)},
\end{matrix}} \right.}\nonumber\\
&{~}{\quad \quad \quad \left. {\begin{matrix}
{{\vec{M}_{{\cal A}{\cal U}[1,2,6:8]  [3:5]}}(k)}\\
{{\vec{M}_{{\cal A}{\cal U}[3:5]  [3:5]}}(k)}
\end{matrix}} \right]},\label{eq:J4_AugMinTorqueSol2_1_a}\\
{\bm{\Xi}_{\cal A}}:=
&{~}{\left[ {\vec{M}_{{\cal A}{\cal U}}^{{\cal A}{\cal A}} - \vec{M}_{{\cal A}{\cal U}}^{{\cal A}{\cal U}}{{\left( {\vec{M}_{{\cal A}{\cal U}}^{{\cal U}{\cal U}}} \right)}^{ - 1}}\vec{M}_{{\cal A}{\cal U}}^{{\cal U}{\cal A}}} \right]^{ - 1}}\nonumber\\
&{~}{\left[ {{{\vec{I}}_{5 \times 5}}, - \vec{M}_{{\cal A}{\cal U}}^{{\cal A}{\cal U}}{{\left( {\vec{M}_{{\cal A}{\cal U}}^{{\cal U}{\cal U}}} \right)}^{ - 1}}} \right]{\begin{bmatrix}
{{\vec{F}_{\cal A}}}\\
{{\vec{F}_{\cal U}}}
\end{bmatrix}}},\label{eq:J4_AugMinTorqueSol2_1_b}\\
{\bm{\Xi}_{\cal U}}: = &{~}{\left[ {\vec{M}_{{\cal A}{\cal U}}^{{\cal U}{\cal U}} - \vec{M}_{{\cal A}{\cal U}}^{{\cal U}{\cal A}}{{\left( {\vec{M}_{{\cal A}{\cal U}}^{{\cal A}{\cal A}}} \right)}^{ - 1}}\vec{M}_{{\cal A}{\cal U}}^{{\cal A}{\cal U}}} \right]^{ - 1}}\nonumber\\
&{~}{\left[ { - \vec{M}_{{\cal A}{\cal U}}^{{\cal U}{\cal A}}{{\left( {\vec{M}_{{\cal A}{\cal U}}^{{\cal A}{\cal A}}} \right)}^{ - 1}},{{\vec{I}}_{3 \times 3}}} \right]{\begin{bmatrix}
{{\vec{F}_{\cal A}}}\\
{{\vec{F}_{\cal U}}}
\end{bmatrix}}}, \label{eq:J4_AugMinTorqueSol2_1_c}
\end{align}
\end{subequations}
where \eqref{eq:J4_AugMinTorqueSol2} is a complete expression that can solve the redundancy resolution problem (using \eqref{eq:J4_AugMinTorqueSol2_a} for actuated joints) as well as disturbance rejection (using \eqref{eq:J4_AugMinTorqueSol2_b} for unactuated joints).

%--------------------------------------------------
% 
%--------------------------------------------------
\subsection{Algorithms of the TOAJ and TOAUJ}
\label{subsec:J4_RedundancyResAlgo}
To implement the formulas (TOAJ and TOAUJ) in \autoref{subsec:J4_MinimumTorqueNorm} and \autoref{subsec:J4_AugmentedMinTorque}, here, the corresponding algorithms (shown in \autoref{algorithm:J4_RedundResAlgo}) are proposed as follows: first, two desired end-effector points (positions ${\vec{p}_{er(i-1)}} \to {\vec{p}_{eri}}$ and velocities ${\dot {\vec{p}}_{er(i-1)}} \to {\dot {\vec{p}}_{eri}}$) and joint-space disturbances ${\bm{\tau}_d}$ are given. The goal is to solve the redundancy resolution problem (i.e., generate an array of actuated joint-space positions/angles and velocities ${{\vec{q}}_{\cal A}}(1:N+1),{\dot {\vec{q}}_{\cal A}}(1:N+1)$) and reject disturbances. In Step{~}\ref{algorithm:J4_RedundResAlgo_state_a} and Step{~}\ref{algorithm:J4_RedundResAlgo_state_b}, the trajectories of $\dot {\vec{p}}_e(k)$ and $\ddot {\vec{p}}_e(k)$ can be generated using different methods, e.g.,{~}\eqref{eq:J4_p2ptraj} in \autoref{sec:J4_NumericalResults}. After completing the iteration, finally, ${{\vec{q}}_{\cal A}}(1:N+1),{\dot {\vec{q}}_{\cal A}}(1:N+1)$ are obtained. The generated data also provides reference states for online control. Additionally, in Step{~}\ref{algorithm:J4_RedundResAlgo_state_c}, for the TOAJ, let ${\vec{K}_{dp \cal U}}$ be ${\bf{0}}$ (indicating {TOAJ} mode) to compare the performance of TOAJ and TOAUJ.

\begin{algorithm*}[htb]
\caption {TOAJ and TOAUJ implementation to solve the redundancy resolution problem via joint-space torque optimization.}
\label{algorithm:J4_RedundResAlgo}
\begin{algorithmic}[1]
\Require {${\vec{p}_{er(i - 1)}},{\dot {\vec{p}}_{er(i - 1)}},{\vec{p}_{eri}},{\dot {\vec{p}}_{eri}}$.}
\Ensure {${{\vec{q}}_{\cal A}}(1:N+1),{\dot {\vec{q}}_{\cal A}}(1:N+1)$.}

\State Initialize {$\vec{q}(1),\dot {\vec{q}}(1),{\vec{p}_e}(1) = {\vec{p}_{er(i - 1)}},{\dot {\vec{p}}_e}(1) = {\dot {\vec{p}}_{er(i - 1)}},{\vec{p}_e}(N + 1) = {\vec{p}_{eri}},{\dot {\vec{p}}_e}(N + 1) = {\dot {\vec{p}}_{eri}}$;}

\For{$k = 1$ to $N + 1$}
\Comment{Trajectory generation from point ${\rm{(}}{\vec{p}_{er(i - 1)}}{\rm{, }}{{\dot {\vec{p}}}_{er(i - 1)}}{\rm{)}}$ to point ${\rm{(}}{\vec{p}_{eri}},{{\dot {\vec{p}}}_{eri}})$.}
\State {$\dot {\vec{p}}_e(k) = {g_v}({\vec{p}_{eri}},{\vec{p}_e}(k)),{\vec{p}_e}(1) = {\vec{p}_{er(i - 1)}}$;}\label{algorithm:J4_RedundResAlgo_state_a} \Comment{${g_v}({\vec{p}_{eri}},{\vec{p}_e}(k))$ is a velocity function, e.g.,{~}\eqref{eq:J4_p2ptraj}.}
\State {$\ddot {\vec{p}}_e(k) = {g_a}({{\dot {\vec{p}}}_{eri}},{{\dot {\vec{p}}}_e}(k)),{{\dot {\vec{p}}}_e}(1) = {{\dot {\vec{p}}}_{er(i - 1)}}$;}\label{algorithm:J4_RedundResAlgo_state_b} \Comment{${g_a}({{\dot {\vec{p}}}_{eri}},{{\dot {\vec{p}}}_e}(k))$ is an acceleration function, e.g.,{~}\eqref{eq:J4_p2ptraj}.}

\State {Compute ${[{\vec{M}}(\vec{q}(k))]_{11 \times 11}},{[{\vec{C}}(\vec{q}(k),\dot {\vec{q}}(k - 1))]_{11 \times 11}},{[{\vec{G}}(\vec{q}(k))]_{11 \times 1}}$ using \eqref{eq:J4_11dof_fwdyn};

$\to {\vec{M}_{{\cal A}{\cal U}}} = {[{\vec{M}}(\vec{q}(k))]_{[1:5,9:11]  [1:5,9:11]}},{[{\vec{C}}(\vec{q}(k),\dot {\vec{q}}(k - 1))]_{[1:5,9:11]  [1:5,9:11]}},{[\vec{G}(\vec{q}(k))]_{[1:5,9:11]}}$ in \eqref{eq:J4_AugMinTorqueSol2_1_a};}

\Comment{\parbox[t]{.5\linewidth}
{${{\vec{q}}_{\cal A}}={\rm{ }}[{p_{mx}},{p_{my}},{\theta_{a1}},{\theta_{a2}},{\theta_{a3}}]_{5 \times 1}^T, {{\vec{q}}_{\cal U}}=[{p_{mz}},{\alpha_m},{\beta_m}]_{3 \times 1}^T$,\\
$\vec{q} = [{p_{mx}},{p_{my}},{p_{mz}},{\alpha_m},{\beta_m},0,{\theta_{p1}},{\theta_{p2}},{\theta_{a1}},{\theta_{a2}},{\theta_{a3}}]_{11 \times 1}^T$.}}

\State {Calculate ${\vec{J}_e}({{\vec{q}}_{\cal A}}(k)),\vec{J}_e^ + ({{\vec{q}}_{\cal A}}(k))$ using \eqref{eq:J4_11dof_fwdyn};}

\State {${{\vec{A}_m}}T \leftarrow ({p_{mx}}(k),{p_{my}}(k))$ using \eqref{eq:J4_4CableOptTension};} %\Comment{Using cable stiffness optimization.}

\State {${{\dot {\vec{q}}}_{\cal A}}(k) \leftarrow {{\dot {\vec{p}}}_e}(k),{{\dot {\vec{q}}}_{\cal A}}(k - 1)$ using \eqref{eq:J4_AugMinTorqueSol2_a};}

\State {${{\dot {\vec{q}}}_{\cal U}}(k) \leftarrow {{\dot {\vec{q}}}_{\cal U}}(k - 1)$ using \eqref{eq:J4_AugMinTorqueSol2_b};}\label{algorithm:J4_RedundResAlgo_state_c} \Comment{For the TOAJ, let ${\vec{K}_{dp \cal U}} = {\bf{0}}$.}

\State {${\theta_{p1}}(k + 1),{\theta_{p2}}(k + 1) \leftarrow {\theta_{a1}}(k),{\theta_{a2}}(k),{\theta_{a3}}(k)$ using \eqref{eq:J4_PenduNominalAngle};} \Comment{Compute the nominal angles of the pendulums.}

\State {$\begin{cases}
\Lambda  = \frac{1}{2}\left\| {\vec{M}_{{\cal A}{\cal U}}^{ - 1}{{[\bm{\tau}_{\cal A}^T,\bm{\tau}_{\cal U}^T]}^T}} \right\|_2^2 \leftarrow \\
\begin{cases}
{{\ddot {\vec{q}}}_{\cal A}}(k) = [\vec{J}_e^ + ({{\vec{q}}_{\cal A}}(k))]\left\{ {\ddot {\vec{p}}_e(k) - [{{\dot {\vec{J}}}_e}({{\vec{q}}_{\cal A}}(k),{{\dot {\vec{q}}}_{\cal A}}(k - 1))]{{\dot {\vec{q}}}_{\cal A}}(k - 1)} \right\},\\
{\begin{bmatrix}
{{\bm{\tau}_{\cal A}}}\\
{{\bm{\tau}_{\cal U}}}
\end{bmatrix}} = \vec{M}_{{\cal A}{\cal U}}^{ - 1}
{\begin{bmatrix}
{{{\ddot {\vec{q}}}_{\cal A}}(k)}\\
{{{\ddot {\vec{q}}}_{\cal U}}(k)}
\end{bmatrix}} + 
{\begin{bmatrix}
{{\vec{F}_{\cal A}}}\\
{{\vec{F}_{\cal U}}}
\end{bmatrix}}
{\text{ using \eqref{eq:J4_AugMinTorqueSol1_1} and \eqref{eq:J4_AugMinTorqueSol2}};}
\end{cases} 
\end{cases}$} % \Comment{Suppose ${{\ddot {\vec{q}}}_{\cal U}}(k) = {\bf{0}}$ for simplification.} %\Comment{To compare TOAJ and TOAUJ performance.}

\State {$\begin{cases}
{{\vec{q}}_{\cal A}}(k + 1) = {{\vec{q}}_{\cal A}}(k) + {{\dot {\vec{q}}}_{\cal A}}(k){T_s}, \\
{{\vec{q}}_{\cal U}}(k + 1) = {{\vec{q}}_{\cal U}}(k) + {{\dot {\vec{q}}}_{\cal U}}(k){T_s};
\end{cases}$} \Comment{Discrete integration.}

\State {${( \cdot )_j} = \begin{cases}
{( \cdot )_{j\max }},\quad {\rm{if}}\;{( \cdot )_j} \ge {( \cdot )_{j\max }},{~}j \in [1,11],\\
{( \cdot )_{j\min }},\quad {\rm{if}}\;{( \cdot )_j} \le {( \cdot )_{j\min }},{~}j \in [1,11],
\end{cases}$}\; {$(\cdot): = \vec{q}(k+1),{~}\dot {\vec{q}}(k+1),{~}{\rm{or}}{~}\delta \dot {\vec{q}}(k+1) = \dot {\vec{q}}(k+1) - \dot {\vec{q}}(k)$;} %\Comment{Saturation constraints.}

\State {${\vec{p}_e}(k + 1) \leftarrow {{\vec{q}}_{\cal A}}(k + 1)$;} \Comment{The computation of ${\vec{p}_e}$ is shown in Appendix{~}\ref{sec:J4_appendix_lagrangian_terms}.}
\If {${\left\| {\vec{p}_{eri}} - {{\vec{p}_e}(k + 1)} \right\|} \leq {\varepsilon}_{\cal A}$} \Comment{Error ${\varepsilon}_{\cal A} > 0$.}
    \State {$i = i + 1$;}
    \Comment{Switch to the next planning point $({\vec{p}_{er(i + 1)}},{{\dot {\vec{p}}}_{er(i + 1)}})$.}
\EndIf
\EndFor
\State {\bf Return} {${{\vec{q}}_{\cal A}}(1:N+1),{\dot {\vec{q}}_{\cal A}}(1:N+1)$.}
\end{algorithmic}
\end{algorithm*}

%==================================================
% \section{Control Design}
%==================================================
\section{Controller Design} \label{sec:J4_OnlineControl}
The joint-space states $({{\vec{q}}_{\cal A}},{\dot {\vec{q}}_{\cal A}},{\theta_{p1}},{\theta_{p2}})$ can be obtained by using~\autoref{algorithm:J4_RedundResAlgo}. One of the advantages of this process is that the state data can be generated offline to reduce the computation cost, particularly, on embedded systems. {\color{black}We can apply the obtained data as joint reference for controller design to extend the study of~\autoref{algorithm:J4_RedundResAlgo} and validation.} The control objective is to track reference trajectories. To achieve this goal, {\color{black}some candidate control strategies such as admittance-based controller~\cite{W.He2020}, adaptive fuzzy controller~\cite{L.Kong2019}, asymmetric bounded neural controller~\cite{L.Kong20192}, etc. may be available; however, in this paper, we are interested in developing a simple yet effective controller.} The proposed nonlinear control law $\vec{u}$ is designed as
\begin{align}
\vec{u}={}& {\rm{diag}}(\vec{A}_{m[1,2][3,4]}^{-1} ,{{\vec{I}}_{2 \times 2}},{{\vec{I}}_{3 \times 3}})\Big\{{\vec{K}_{{\cal A}p}}{{\vec{e}}_{\cal A}} + {\vec{K}_{{\cal A}d}}{\dot {\vec{e}}_{\cal A}}\nonumber\\
{}& {+{\vec{K}_{{\cal A}i}}\int_0^t {{{\vec{e}}_{\cal A}}dt}} \Big\},
\label{eq:J4_PID}
\end{align}
where ${\vec{K}_{{\cal A}p}}$, ${\vec{K}_{{\cal A}d}}$, and ${\vec{K}_{{\cal A}i}}$ are positive gains. The error vector between the reference and measured values is defined as ${\vec{e}}_{\cal A}=[{p_{mx}},{p_{my}},{\theta_{p1}},{\theta_{p2}},{\theta_{a1}},{\theta_{a2}},{\theta_{a3}}{]^T}-[{\hat p_{mx}},{\hat p_{my}},{\hat \theta_{p1}},{\hat \theta_{p2}},{\hat \theta_{a1}},{\hat \theta_{a2}},{\hat \theta_{a3}}{]^T}.$ ${\dot {\vec{e}}_{\cal A}}$ is the time-derivative of ${\vec{e}}_{\cal A}$ denoting velocity errors. Regarding \autoref{subsec:J4_ModelReduction}, one can also get $\vec{u}={[{\delta T_3},{\delta T_4},{\tau_{p1}},{\tau_{p2}},{\tau_{a1}},{\tau_{a2}},{\tau_{a3}}]^T}$, so the controller \eqref{eq:J4_PID} has the following abilities: damp in-plane vibrations using lower cable tensions $({\delta T_3},{\delta T_4})$, eliminate out-of-plane motions of the platform ${{\vec{q}}_{\cal U}}$ using the torques of two pendulums $({\tau_{p1}},{\tau_{p2}})$, and generate joint torques $({{\tau_{a1}},{\tau_{a2}},{\tau_{a3}}})$ for the rigid robot arm. In this way, the controller \eqref{eq:J4_PID} attempts to minimize the tracking errors over time by adjusting $\vec{u}$.

%%%%%%%%%%%%%%%%%%%%%%%%%%%%%%%%%%%%%%%%%%%%%%%%%%%%%%%
% Source: J4_05312019
%%%%%%%%%%%%%%%%%%%%%%%%%%%%%%%%%%%%%%%%%%%%%%%%%%%%%%%
\section{Numerical Results}\label{sec:J4_NumericalResults}
To evaluate the performance of \autoref{sec:J4_RedundancyResolution} and \autoref{sec:J4_OnlineControl}, we conduct the following case studies. All the scenarios are implemented using MATLAB 2019a (The MathWorks, Inc.) on a Windows 10 x64 desktop PC (Intel Core i7-2600, 3.4 GHz CPU and 12.0 GB RAM).

\subsection{Scenario 1: Point-to-Point Trajectory}
\label{subsec:J4_p2ptrajplanning}
The end-effector trajectory conducted by a normalized polynomial (here, one of trajectory generation methods and constraints that have been applied in practice in{~}\cite{Flacco2015,Nabat2005} are utilized for case studies) from Cartesian point ${\rm{(}}{\vec{p}_{er(i - 1)}},{{\dot {\vec{p}}}_{er(i - 1)}}{\rm{)}}$ to Cartesian point ${\rm{(}}{\vec{p}_{eri}},{{\dot {\vec{p}}}_{eri}})$ is given as follows:
\begin{align}
\begin{cases}
N = \frac{{{t_i} - {t_{i - 1}}}}{{{t_s}}},\eta  = \frac{k}{N},k \in [1,N+1]\\
{{\bm{\rho}}_r} = {\vec{p}_{er(i - 1)}} + ({\vec{p}_{eri}} - {\vec{p}_{er(i - 1)}})(6{\eta ^5} - 15{\eta ^4} + 10{\eta ^3})\\
{{\dot {\bm{\rho}}}_r} = \frac{{{\vec{p}_{eri}} - {\vec{p}_{er(i - 1)}}}}{{{t_i} - {t_{i - 1}}}}(30{\eta ^4} - 60{\eta ^3} + 30{\eta ^2})\\
{{\ddot {\bm{\rho}}}_r} = \frac{{{\vec{p}_{eri}} - {\vec{p}_{er(i - 1)}}}}{{{t_i} - {t_{i - 1}}}}(120{\eta ^3} - 180{\eta ^2} + 60\eta )\\
{{\dot {\vec{p}}}_e}(k) = {{\dot {\bm{\rho}}}_r} + 10({{\bm{\rho}}_r} - {\vec{p}_e}(k))\\
{{\ddot {\vec{p}}}_e}(k) = {{\ddot {\bm{\rho}}}_r} + 10({{\dot {\bm{\rho}}}_r} - {{\dot {\vec{p}}}_e}(k))
\end{cases},
\label{eq:J4_p2ptraj}
\end{align}
where the start time ${t_{i - 1}}$, end time ${t_i}$, and sampling time ${t_s}$ are supposed to be $0{{~}\rm{s}}$, $1{{~}\rm{s}}$, and $0.0002{{~}\rm{s}}$, respectively. Let the positions and velocities of Cartesian points ${\vec{p}_{er(i - 1)}}=[0,0.334,0]^T {{~}\rm{m}}$, ${\vec{p}_{eri}}=[0.35,0.5,0.1]^T {{~}\rm{m}}$, ${{\dot {\vec{p}}}_{er(i - 1)}}=[0,0,0]^T {{~}\rm{m/s}}$, and ${{\dot {\vec{p}}}_{eri}}=[0,0,0]^T {{~}\rm{m/s}}$, respectively. The damping gain ${\vec{K}_{dp{\cal A}}={{\rm{diag}}(500,500,500,500,500)}}$. Furthermore, the constraints are also given as ${\dot {\vec{q}}_{{\cal A}\rm{max}}}=[3{{~}\rm{m/s}},3{{~}\rm{m/s}},25{{~}\rm{rad/s}},25{{~}\rm{rad/s}},25{{~}\rm{rad/s}}]^T$, ${\dot {\vec{q}}_{{\cal A}\rm{min}}}=-$ $[3{{~}\rm{m/s}},3{{~}\rm{m/s}},25{{~}\rm{rad/s}},25{{~}\rm{rad/s}},25{{~}\rm{rad/s}}]^T$, ${\delta \dot {\vec{q}}_{{\cal A}\rm{max}}} = [30{{~}\rm{m/s}},30{{~}\rm{m/s}},250{{~}\rm{rad/s}},250{{~}\rm{rad/s}},250{{~}\rm{rad/s}}]^T$, and ${\delta \dot {\vec{q}}_{{\cal A}\rm{min}}} = -[30{{~}\rm{m/s}},30{{~}\rm{m/s}},250{{~}\rm{rad/s}},250{{~}\rm{rad/s}},250$ ${\rm{rad/s}}]^T$. There are no constraints to ${\dot {\vec{q}}_{{\cal U}}}$ and ${\delta \dot {\vec{q}}_{{\cal U}}}$. The error ${\vec{\varepsilon}_{\cal A}}$ is set to $\rm{2.22e{-16}}$. Then,{~}\autoref{algorithm:J4_RedundResAlgo} is implemented, and the following performance indices of the TOAJ and TOAUJ are compared.

\begin{figure}[!t]\centering	
	\includegraphics[width=75mm]{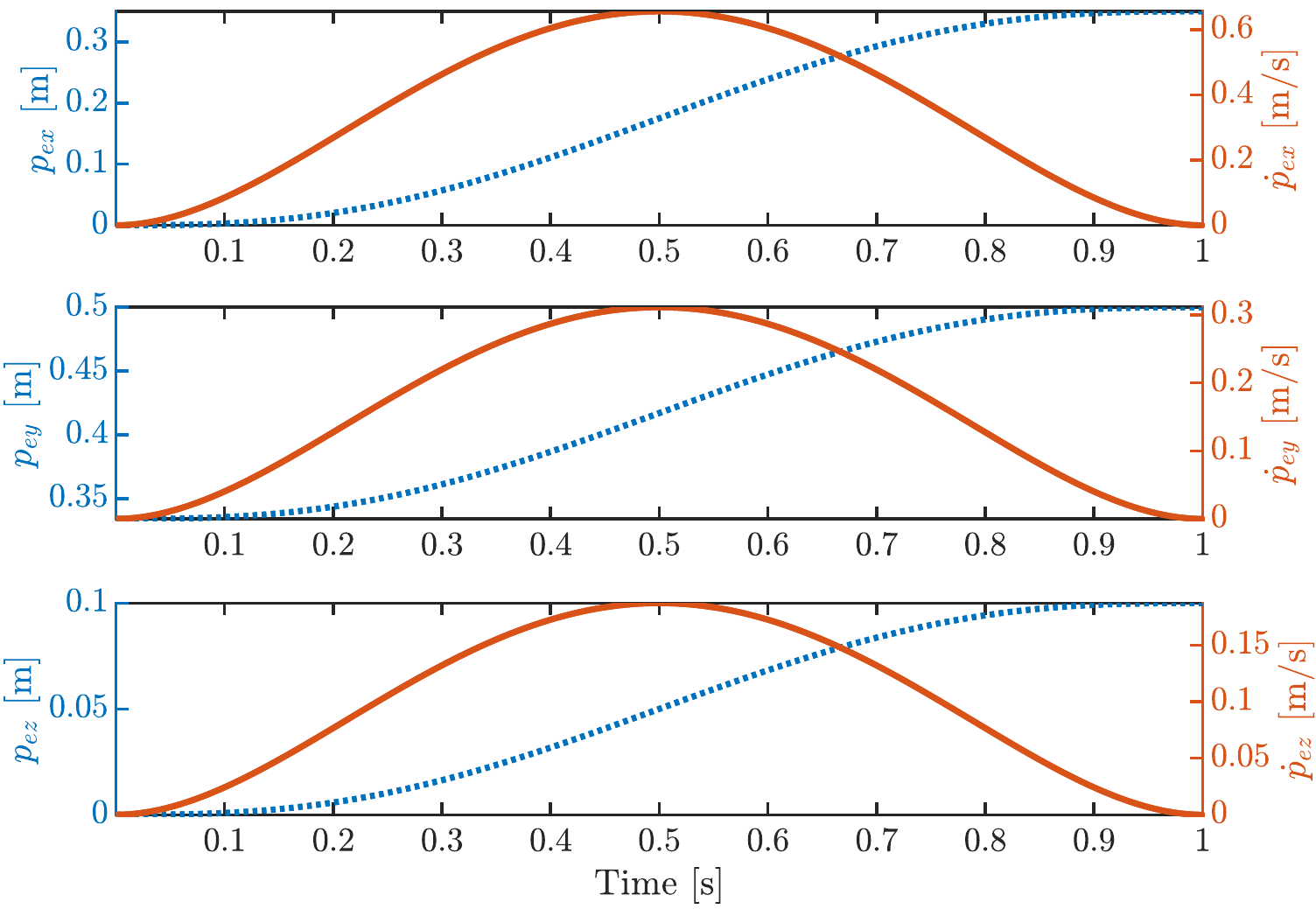}	
	\caption{Cartesian positions and velocities of the end-effector.}
	\label{fig:J4_ReduRes_RefEndEffectorPosVel}
	\vspace{-0.3cm}
\end{figure}

\begin{figure}[!t]\centering	
	\includegraphics[width=75mm]{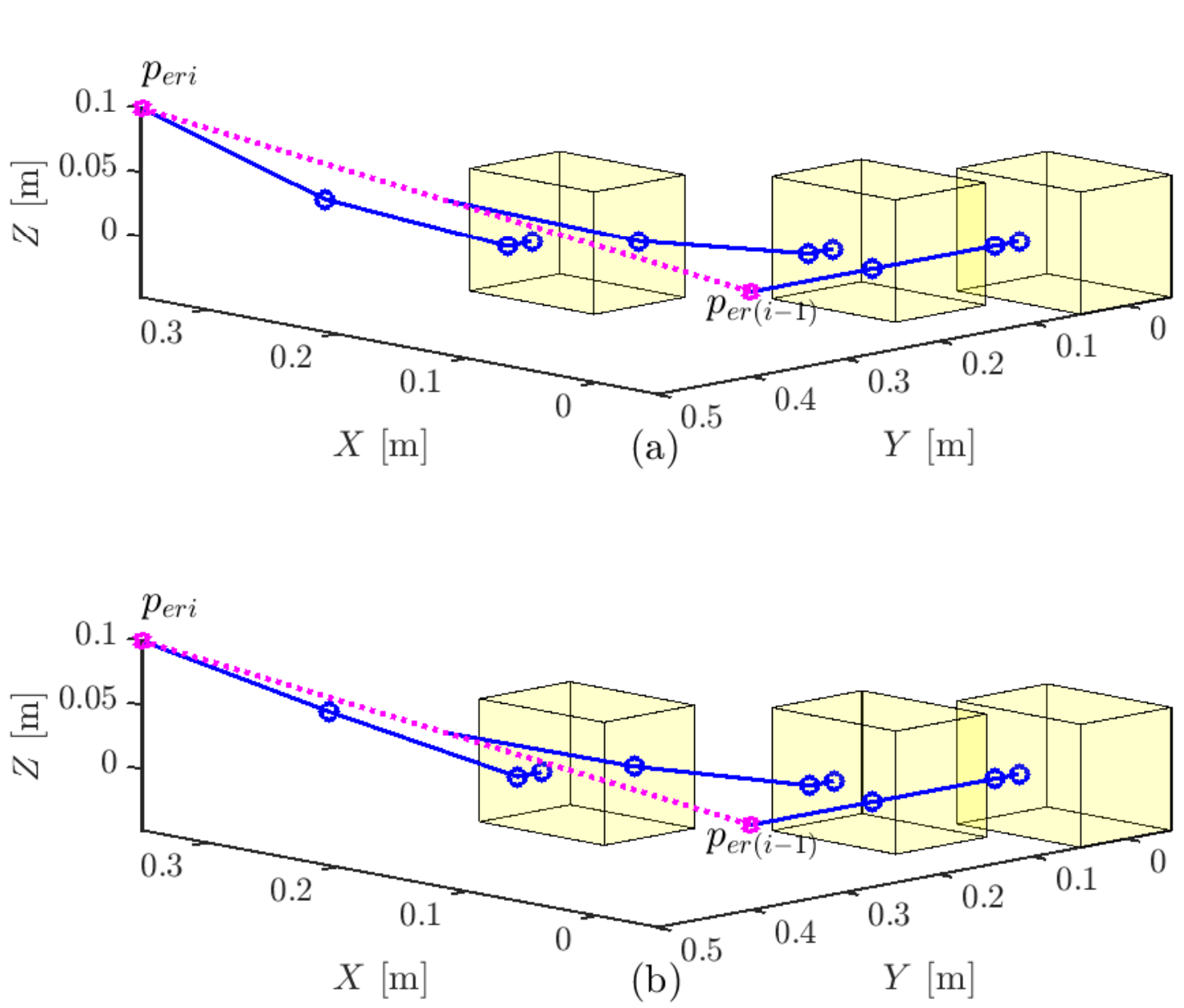}
	\vspace{-0.3cm}
	\caption{Trajectory responses of the HCDR by given the start point	${\vec{p}_{er(i - 1)}}=[0,0.334,0]^T {{~}\rm{m}}$ and the end point ${\vec{p}_{eri}}=[0.35,0.5,0.1]^T {{~}\rm{m}}$, where the \emph{yellow cube}, \emph{blue lines}, \emph{blue circles}, and \emph{magenta dotted line} represent the mobile platform (cables are not displayed here), links of the robot arm, joints of the robot arm, and trajectory of the end-effector, respectively. (a) TOAJ and (b) TOAUJ.}
	\label{fig:J4_ReduRes_RobTrajectory}
    \vspace{-0.3cm}
\end{figure}

\begin{figure}[!t]\centering	
	\includegraphics[width=75mm]{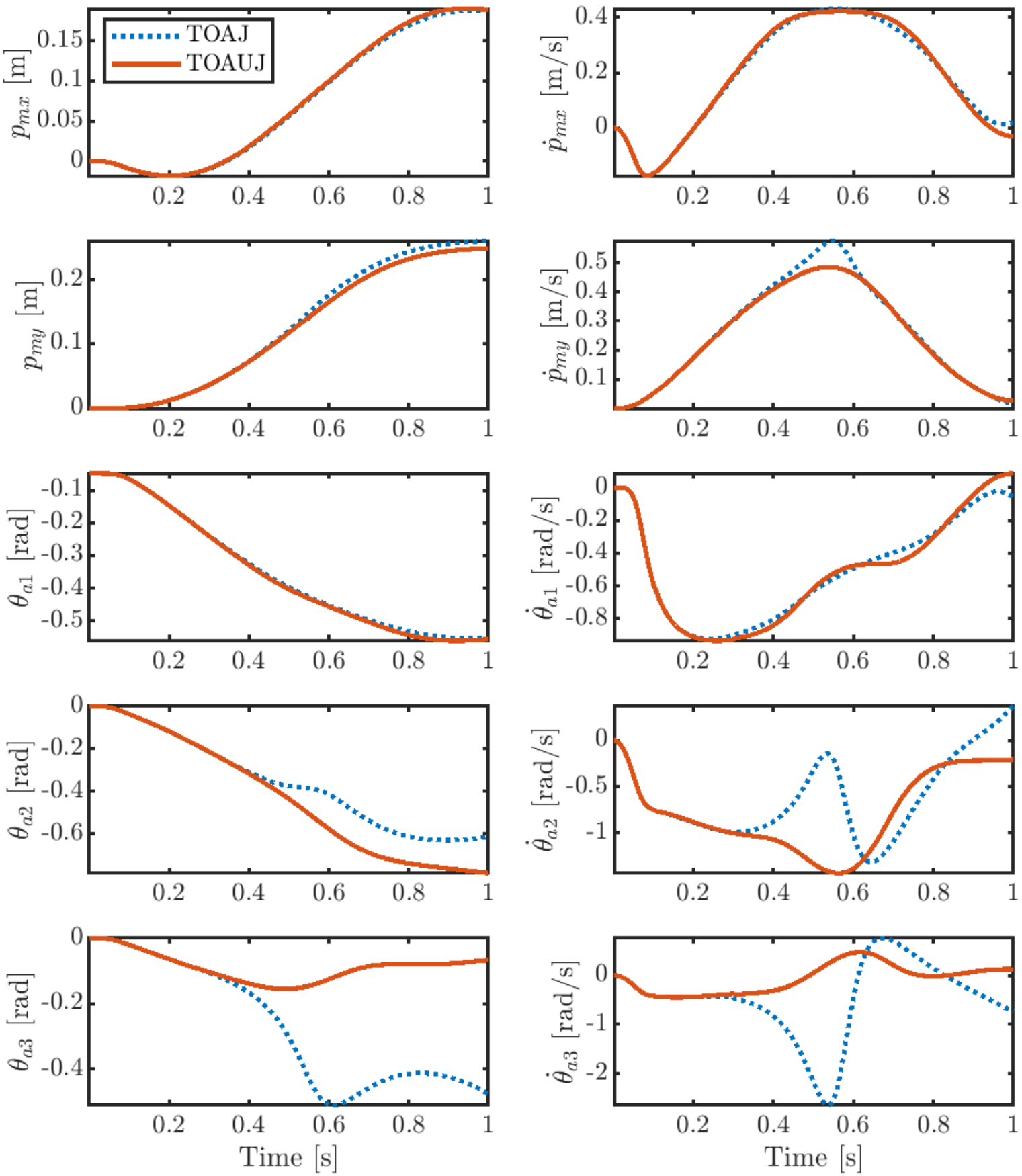}
	\vspace{-0.3cm}
	\caption{Redundancy resolution of the actuated joints.}
	\label{fig:J4_ReduRes_ActuatedJointStates}
	\vspace{-0.3cm}
\end{figure}

The results in \autoref{fig:J4_ReduRes_RefEndEffectorPosVel} and \autoref{fig:J4_ReduRes_RobTrajectory} show Cartesian positions and velocities of the end-effector and trajectory responses of the HCDR by given the start point ${\vec{p}_{er(i-1)}}=[0,0.334,0]^T {{~}\rm{m}}$ and the end point ${\vec{p}_{eri}}=[0.35,0.5,0.1]^T {{~}\rm{m}}$, respectively. The corresponding redundancy resolution of the actuated joints is shown in{~}\autoref{fig:J4_ReduRes_ActuatedJointStates}. Clearly, using TOAJ and TOAUJ one can get smooth trajectory and redundancy resolution responses. Compared to TOAJ, TOAUJ shows a better performance in contributing to avoiding singularity, i.e., ${\theta_{a3}}$ is not close to zero when the HCDR moves to the end point ${\vec{p}_{eri}}$.

\begin{figure}[!t]\centering	
	\includegraphics[width=75mm]{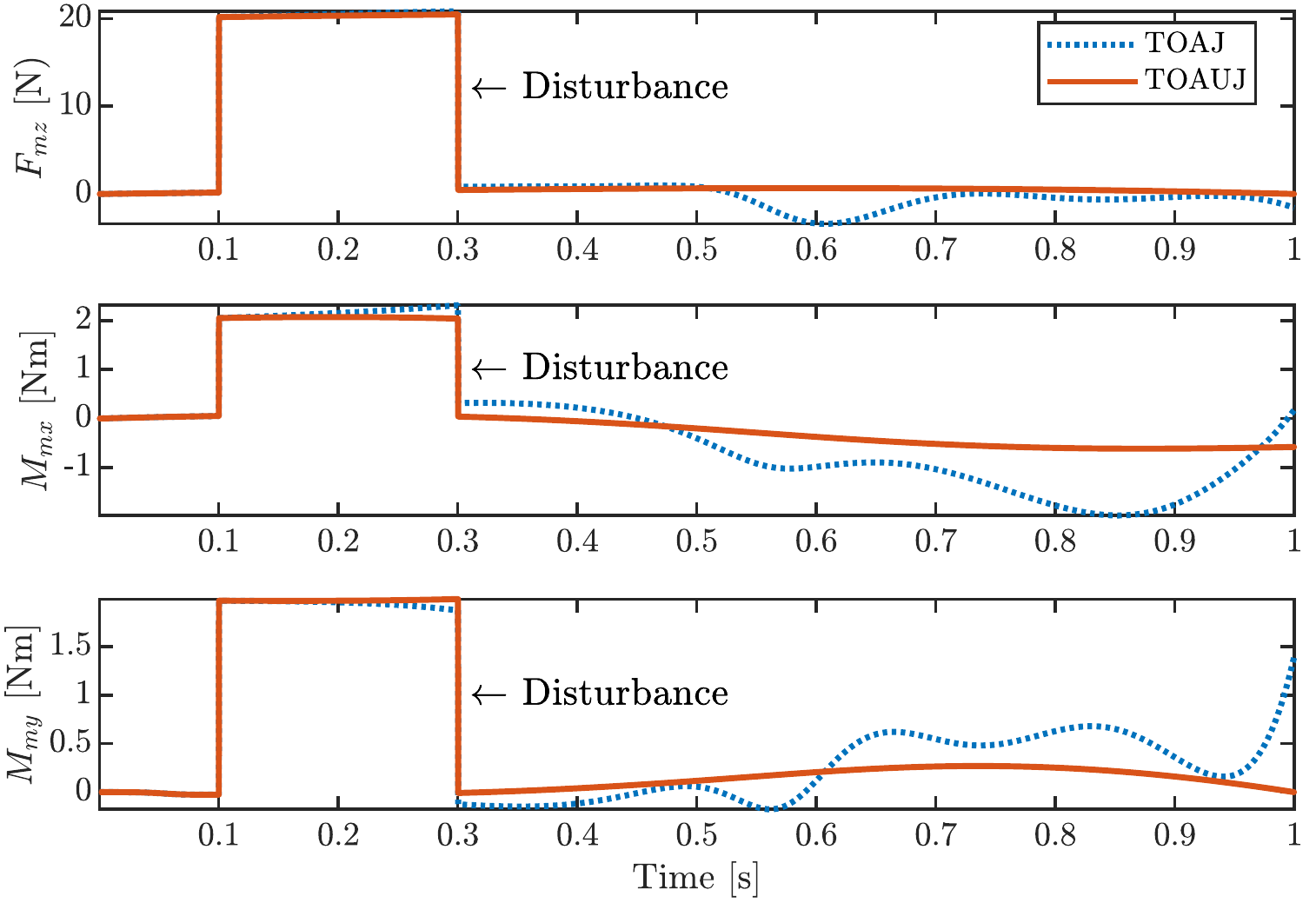}
	\vspace{-0.3cm}
	\caption{Force/torque responses of the unactuated joints.}
	\label{fig:J4_ReduRes_UnactuatedJointTorque}
    \vspace{-0.3cm}
\end{figure}

\begin{figure}[!t]\centering	
	\includegraphics[width=75mm]{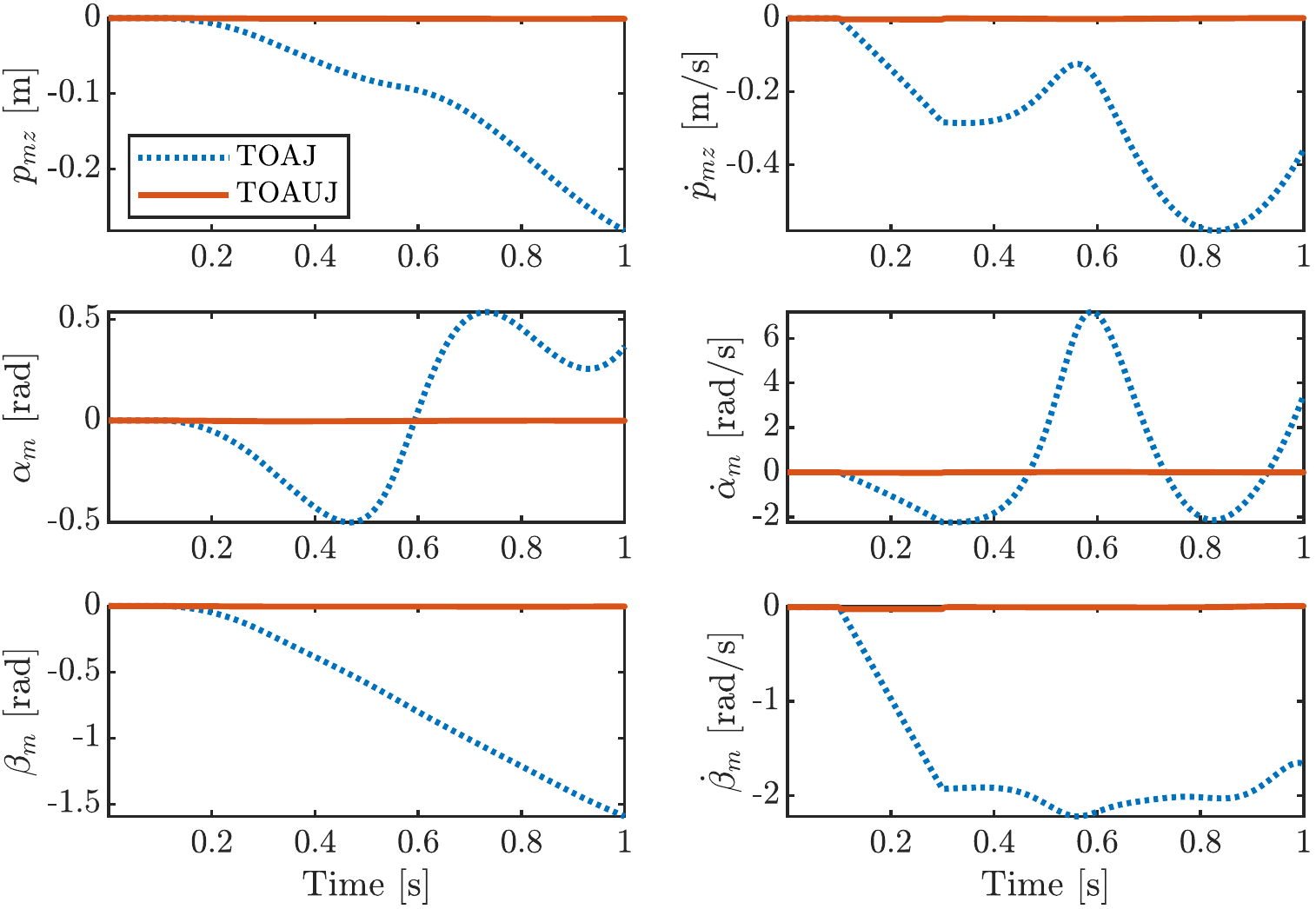} 
	\vspace{-0.3cm}
	\caption{State responses of the unactuated joints.}
	\label{fig:J4_ReduRes_UnactuatedJointStates}
    \vspace{-0.3cm}
\end{figure}

In{~}\autoref{fig:J4_ReduRes_UnactuatedJointTorque}, $20{{~}\rm{N}}$, $2{{~}\rm{Nm}}$, and $2{{~}\rm{Nm}}$ pulse disturbances (from $0.1{{~}\rm{s}}$ to $0.3{{~}\rm{s}}$) are given to three unactuated joints, respectively. The results show that the force/torque responses (${F_{mz}}$, ${M_{mx}}$, and ${M_{my}}$) of the unactuated joints TOAUJ are able to stabilize unactuated joints motions, while TOAJ can not; meanwhile, these results are validated by the state responses of the unactuated joints (shown in~\autoref{fig:J4_ReduRes_UnactuatedJointStates}).

\begin{figure}[!t]\centering	
	\includegraphics[width=85mm]{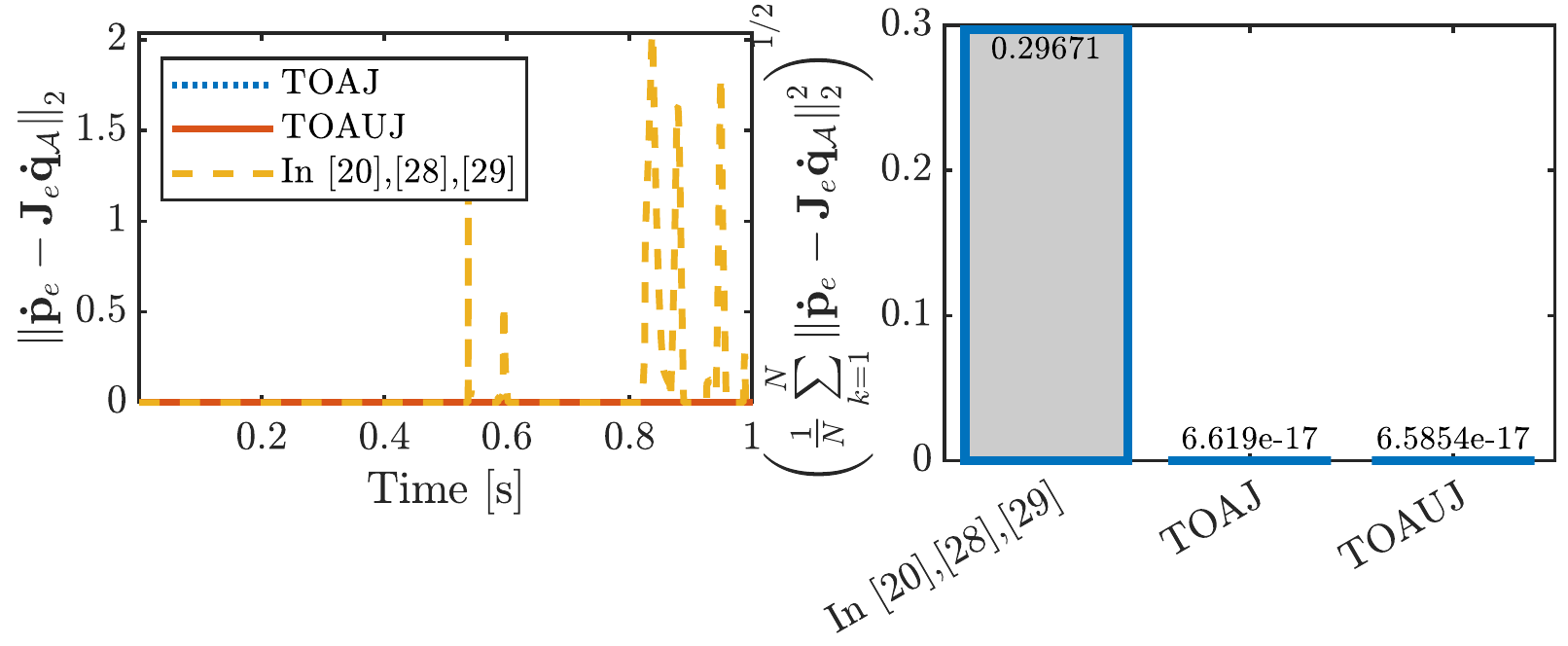}
	\vspace{-0.3cm}
	\caption{\color{black}Error performance by using different methods.}
	\label{fig:J4_ReduRes_NormEndEffectorError_RMSE}
    \vspace{-0.5cm}
\end{figure} 

Additionally, to evaluate the error performance ${\left\| {{{\dot {\vec{p}}}_e} - {\vec{J}_e}{{\dot {\vec{q}}}_{\cal A}}} \right\|_2}$ and ${\left( {\frac{1}{N}\sum\limits_{k = 1}^N {\left\| {{{\dot {\vec{p}}}_e} - {\vec{J}_e}{{\dot {\vec{q}}}_{\cal A}}} \right\|_2^2} } \right)^{1/2}}$ by using different methods (with the same setup), i.e., in comparison with the minimum weighted torque norm in{~}\cite{Kazerounian1988,S.Ma1994,Flacco2015} and the proposed approaches (TOAJ and TOAUJ), and the results are shown in{~}\autoref{fig:J4_ReduRes_NormEndEffectorError_RMSE}. Clearly, the errors via TOAJ and TOAUJ are much less than that of{~}\cite{Kazerounian1988,S.Ma1994,Flacco2015}. The results also verify the effectiveness of the proposed methods. {\color{black}In this scenario, the average running time by using TOAJ and TOAUJ methods is about 77.6 s. Although MATLAB in this paper is not a real-time environment, it validates that the reference trajectories can be generated offline. The generated data can be considered as reference command for motion planning. We can also expect that the real-time performance will be improved on a real-time platform.}

\subsection{Scenario 2: Point-to-Point Trajectory with the Controller}
\label{subsec:J4_p2p_online_ctrl}
\begin{figure}[!t]\centering
    \begin{subfigure}
        \centering
        \includegraphics[width=43mm]{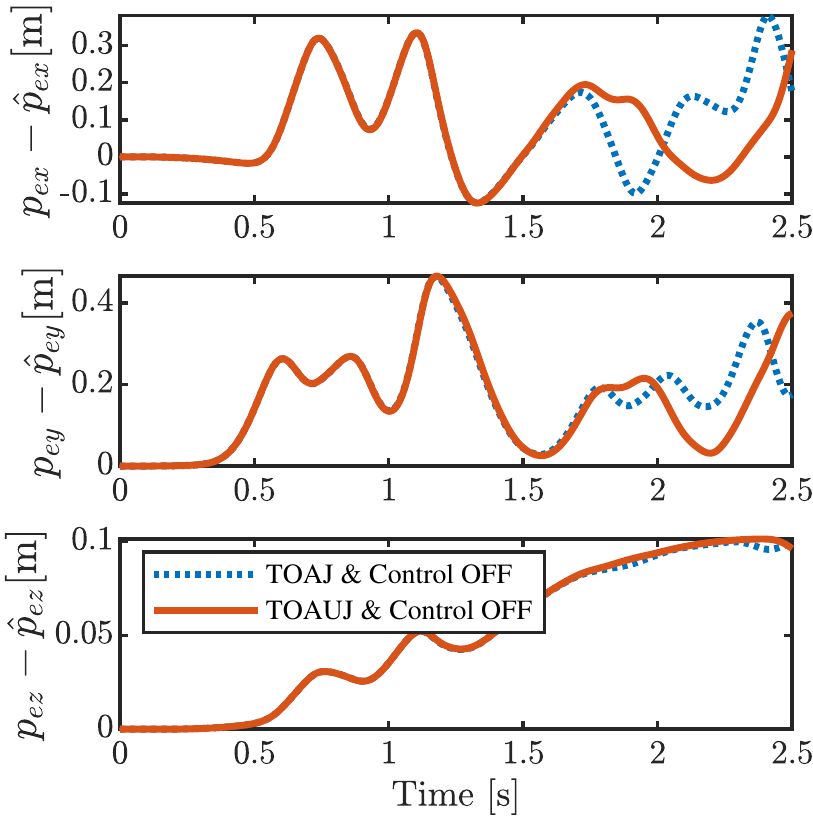}
        \label{fig:J4_CtrlOFF_EndEffector_vs_Time_WithLim}
        \end{subfigure}
    \begin{subfigure}
        \centering
        \includegraphics[width=43mm]{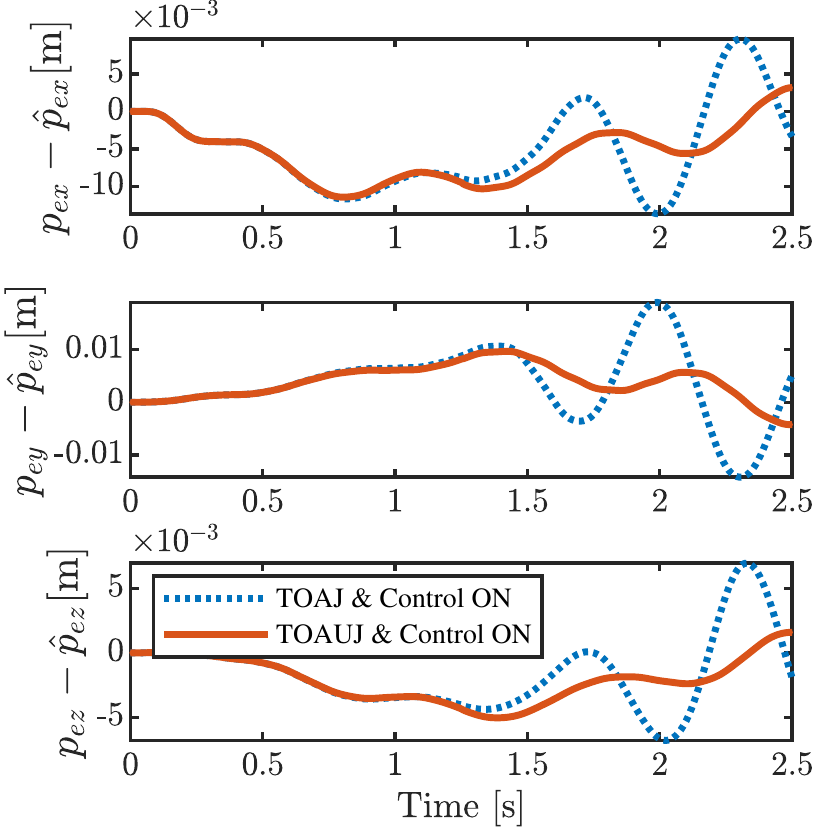}
        \label{fig:J4_CtrlON_EndEffector_vs_Time_WithLim}
    \end{subfigure}
    \vspace{-0.6cm}
    \caption{{\color{black}Position tracking errors of the end-effector by using different control strategies.}}
	\label{fig:J4_Ctrl_EndEffector_vs_Time_WithLim}
\vspace{-0.5cm}
\end{figure}

In order to evaluate the performance of point-to-point trajectory with the controller, we apply the generated joint data (as reference command) in {~}\autoref{fig:J4_ReduRes_ActuatedJointStates} to the controller{~}\eqref{eq:J4_PID}. Here, {\color{black}we provide the position tracking errors of the end-effector by using different methods (see~\autoref{fig:J4_Ctrl_EndEffector_vs_Time_WithLim}) to illustrate the effectiveness of the control law{~}\eqref{eq:J4_PID}, including (a) Control OFF, i.e., ${\vec{K}_{{\cal A}p}}$, ${\vec{K}_{{\cal A}d}}$, and ${\vec{K}_{{\cal A}i}}$ are all equal to zero, and (b) Control ON, i.e., {\color{black}the gains of the controller{~}\eqref{eq:J4_PID} are obtained by sufficiently tuning:} ${\vec{K}_{{\cal A}p}}=\rm{diag}{(1,1,0.001,0.001,1.5,1.8,1.5)}$, ${\vec{K}_{{\cal A}d}}=\rm{diag}{(10,30,0.1,0.1,0.05,0.09,0.05)}$, and ${\vec{K}_{{\cal A}i}}=\rm{diag}{(1,1,0.1,0.1,2,6.75,5)}$. We can summarize the following results from~\autoref{fig:J4_Ctrl_EndEffector_vs_Time_WithLim}: when the controller is off, the maximum tracking error between the reference position $({p_{ex}}, {p_{ey}}, {p_{ez}})$ (given joint command to find end-effector position by using{~}\eqref{eq:J4_appendix_Pe}) and the measured position $({\hat p_{ex}}, {\hat p_{ey}}, {\hat p_{ez}})$ is $(0.381, 0.466, 0.101) \;\rm{m}$; when the controller is on, the maximum tracking error is reduced to $(0.014, 0.019, 0.007) \;\rm{m}$. Clearly, the control law~\eqref{eq:J4_PID} is effective, and TOAUJ-based method holds a better tracking performance than that of TOAJ. Meanwhile, the controller can also help position control by collaborating with the TOAJ-based and TOAUJ-based approaches. Compared to the existing control methods such as in~\cite{W.He2020,L.Kong2019,L.Kong20192}, Eq.~\eqref{eq:J4_PID} is a simpler yet effective control strategy. It is easier to implement in practice, and the given trajectory can also be tracked in real-time.

Additionally, this paper provides a more general configuration (i.e., more DOFs in the 3D) of HCDR for system modeling and algorithms developing, which can cover fewer DOF configurations, such as a 2D HCDR in~\cite{R.Qi2019j4}, indicating that the validated simulations in this paper can also be applied to such HCDRs.}

%==================================================
% \section{Conclusions and Future Work}
%==================================================
\section{Conclusions}
\label{sec:J4_Conclusions}
This paper proposed two new methods to solve redundancy resolution based on the reduced dynamic model, in which the first method called TOAJ. The basic idea of this method was similar to the existing approaches, but we introduced a damping gain to stabilize self-motions. Though the TOAJ approach, we could obtain the redundancy resolution acting on actuated joints. The second approach called TOAUJ; using this method, we could not only get the redundancy resolution for actuated joints but also reject disturbances appearing in unactuated joints. Besides, we also provided detailed algorithms to implement TOAJ and TOAUJ. Numerical results suggested the effectiveness of these two methods. {\color{black}In the future, we plan to design a real HCDR system that can hold the implementation of TOAJ and TOAUJ simultaneously for experiments (e.g., warehousing and rehabilitation applications).
}

%==================================================
% \appendices
%==================================================
\appendices
\section{}
\label{sec:J4_appendix_lagrangian_terms}
The corresponding terms in \eqref{eq:J4_kinetic_energy} and \eqref{eq:J4_potential_energy} are computed as follows: the rotation matrix $\vec{R}_g^m$ is equal to $\vec{R}_x({\alpha_m})\vec{R}_{y'}({\beta_m})\vec{R}_{z''}({\gamma_m}) \in {\mathcal{SO}(3)}$ and the angular velocity ${{{\bm{\omega}}}_m}$ of the frame $\{O_m\}$ is calculated as
{\color{black}
    \begin{align}
        \bm{\omega}_m = &{\;}[\vec{R}_g^m]^T[{{\dot \alpha}_m},0,0]^T + [\vec{R}_{m\beta }^{m\gamma }]^T[\vec{R}_{m\alpha }^{m\beta }]^T {[0,{{\dot \beta }_m},0]^T}  \nonumber\\
        &{\;}+ {[\vec{R}_{m\beta }^{m\gamma }]^T}[0,0,{{\dot \gamma }_m}]^T,
    \label{eq:J4_appendix_Wm}
\end{align}
with the rotation matrices $\vec{R}_{m\alpha}^{m\beta}$ $=$ $\vec{R}_{y'}({\beta_m})$ $\in$ $ {\mathcal{SO}(3)}$ and $\vec{R}_{m\beta}^{m\gamma}$ $=$ $\vec{R}_{z''}({\gamma_m})$ $\in$ $ {\mathcal{SO}(3)}$. Then, the COM (of the pendulums' links) positions are computed as
\begin{subequations}
    \label{eq:J4_appendix_Ppc}
    \begin{align}
        \vec{p}_{pc1} = &{\;}\vec{p}_{p1_0}+\vec{R}_g^m{\vec{R}_x}({\theta_{p1}})[x_{pc01},y_{pc01},z_{pc01}]^T,
        \label{eq:J4_appendix_Ppc2}\\
        \vec{p}_{pc2} = &{\;}{\vec{p}_{p2_0}}+\vec{R}_g^m{\vec{R}_x}({\theta_{p2}})[x_{pc02},y_{pc02},z_{pc02}]^T,
        \label{eq:J4_appendix_Ppc2}
    \end{align}
\end{subequations}
where the joint position vectors are described as 
\begin{subequations}
    \label{eq:J4_appendix_Pp}
    \begin{align}
        \vec{p}_{p1_0} = &{\;}[{p_{mx}},{p_{my}},{p_{mz}}]^T +  \vec{R}_g^m[x_{p01},y_{p01},z_{p01}]^T,
        \label{eq:J4_appendix_Pp10}\\ 
        \vec{p}_{p2_0} = &{\;}[{p_{mx}},{p_{my}},{p_{mz}}]^T + \vec{R}_g^m[x_{p02},y_{p02},z_{p02}]^T.
        \label{eq:J4_appendix_Pp20}
    \end{align}
\end{subequations}
The COM linear velocities and angle velocities (of the pendulums' links) are calculated as 
\begin{subequations}
    \label{eq:J4_appendix_VpcWpc}
    \begin{align}
        {\vec{v}_{pc1}} = &{\;} {\dot {\vec{p}}}_{pc1}, 
        \label{eq:J4_appendix_Vpc1}\\
        {\vec{v}_{pc2}} = &{\;} {\dot {\vec{p}}}_{pc2},
        \label{eq:J4_appendix_Vpc2}\\ 
        {\bm{\omega}_{pc1}} = &{\;} ({\vec{R}_x}({\theta_{p1}}))^T{\bm{\omega}_m} + [{{\dot \theta }_{p1}},0,0]^T,
        \label{eq:J4_appendix_Wpc1}\\
        {\bm{\omega}_{pc2}} = &{\;} ({\vec{R}_x}({\theta_{p2}}))^T{\bm{\omega}_m} + [{{\dot \theta }_{p2}},0,0]^T,
        \label{eq:J4_appendix_Wpc2}
    \end{align}
\end{subequations}
with the corresponding parameters are shown in~\autoref{fig:J4_11DOFHCDRModel} and~\autoref{table:J4_11DOFHCDRParameters}. Then, the COM (of the links) positions are computed as 
\begin{subequations}
    \label{eq:J4_appendix_Pac}
    \begin{align}
        {\vec{p}_{ac1}} =&{\;} {\vec{p}_{a0}} + \vec{R}_g^m{\vec{R}_{y}}({\theta_{a1}})[x_{ac01},y_{ac01},z_{ac01}]^T, 
        \label{eq:J4_appendix_Pac1}\\
        {\vec{p}_{ac2}} =&{\;} {\vec{p}_{a1}} + \vec{R}_g^m{\vec{R}_{y}}({\theta_{a1}}){\vec{R}_{z}}({\theta_{a2}})[x_{ac02},y_{ac02},z_{ac02}]^T, \label{eq:J4_appendix_Pac2}\\
        {\vec{p}_{ac3}} =&{\;} {\vec{p}_{a2}} + \vec{R}_g^m{\vec{R}_{y}}({\theta_{a1}}){\vec{R}_{z}}({\theta_{a2}}){\vec{R}_{z}}({\theta_{a3}})\nonumber\\
        &{\;}[x_{ac03},y_{ac03},z_{ac03}]^T,
        \label{eq:J4_appendix_Pac3}
    \end{align}
\end{subequations}
where the joint position vectors are calculated as
\begin{subequations}
    \label{eq:J4_appendix_Pa}
    \begin{align}
        {\vec{p}_{a0}} = &{\;}[{p_{mx}},{p_{my}},{p_{mz}}]^T + \vec{R}_g^m[x_{a0},y_{a0},z_{a0}]^T,
        \label{eq:J4_appendix_Pa1}\\
        {\vec{p}_{a1}} = &{\;}{\vec{p}_{a0}} + \vec{R}_g^m{\vec{R}_{y}}({\theta_{a1}})[x_{a01},y_{a01},z_{a01}]^T,
        \label{eq:J4_appendix_Pa2}\\
        {\vec{p}_{a2}} = &{\;}{\vec{p}_{a1}}  +  \vec{R}_g^m{\vec{R}_{y}}({\theta_{a1}}){\vec{R}_{z}}({\theta_{a2}})[x_{a02},y_{a02},z_{a02}]^T,
        \label{eq:J4_appendix_Pa3}\\
        {\vec{p}_e} = &{\;}{\vec{p}_{a3}} = {\vec{p}_{a2}}  +  \vec{R}_g^m{\vec{R}_{y}}({\theta_{a1}}){\vec{R}_{z}}({\theta_{a2}}){\vec{R}_{z}}({\theta_{a3}})\nonumber\\
        &{\;}[x_{a03},y_{a03},z_{a03}]^T.
        \label{eq:J4_appendix_Pe}
    \end{align}
\end{subequations}
Additionally, the COM linear velocities and angle velocities (of the links) are calculated as 
\begin{subequations}
    \label{eq:J4_appendix_VacWac}
    \begin{align}
        {\vec{v}_{ac1}} = &{\;} {\dot {\vec{p}}}_{ac1},
        \label{eq:J4_appendix_Vac1}\\ 
        {\vec{v}_{ac2}} = &{\;} {\dot {\vec{p}}}_{ac2},
        \label{eq:J4_appendix_Vac2}\\ 
        {\vec{v}_{ac3}} = &{\;} {\dot {\vec{p}}}_{ac3},
        \label{eq:J4_appendix_Vac3}\\
        {\bm{\omega}_{ac1}} = &{\;} ({\vec{R}_{y}}({\theta_{a1}}))^T{\bm{\omega}_m}  +  [0,{{{\dot \theta }_{a1}},0}]^T,
        \label{eq:J4_appendix_Wac1}\\
        {\bm{\omega}_{ac2}} = &{\;} ({\vec{R}_{y}}({\theta_{a1}}){\vec{R}_{z}}({\theta_{a2}}))^T({\bm{\omega}_m}  + [0,{{{\dot \theta }_{a1}},0}]^T) \nonumber\\
        &{\;} + [0,{0,{{\dot \theta }_{a2}}}]^T,
        \label{eq:J4_appendix_Wac2}\\
        {\bm{\omega}_{ac3}} = &{\;} ({\vec{R}_{y}}({\theta_{a1}}){\vec{R}_{z}}({\theta_{a2}}){\vec{R}_{z}}({\theta_{a3}}))^T({\bm{\omega}_m}  + [0,{{{\dot \theta }_{a1}},0}]^T) +\nonumber\\ &{\;}({\vec{R}_{z}}({\theta_{a2}}){\vec{R}_{z}}({\theta_{a3}}))^T[0,0,{{\dot \theta }_{a2}}]^T + [0,{0,{{\dot \theta }_{a3}}}]^T,
        \label{eq:J4_appendix_Wac3}
    \end{align}
\end{subequations}
where the corresponding parameters are also shown in~\autoref{fig:J4_11DOFHCDRModel} and \autoref{table:J4_11DOFHCDRParameters}.

In \eqref{eq:J4_potential_energy}, the cable stiffness matrix ${\vec{K}_c}$ equals ${\rm{diag}}$ $\left( {{\frac{EA_1}{L_{01}}},{\frac{EA_2}{L_{02}}}, \cdots ,{\frac{EA_{12}}{L_{012}}}} \right) $ $\in$ $ {\mathbb{R}^{12 \times 12}}$, where ${\frac{EA_i}{L_{0i}}}$ represents the $i$th cable stiffness, $EA_i$ is the product of the modulus of elasticity and cross-sectional area  of the $i$th cable. $L_{0i}$ and $L_{i}$ denote the $i$th input unstretched cable length and actual cable length, respectively. Vectors ${\vec{L}}_0$ and ${\vec{L}}$ are equal to $[L_{01},L_{02},\cdots,L_{012}]^T$ $\in$ $ {\mathbb{R}^{12}}$ and $[L_1,L_2,\cdots,L_{12}]^T$ $\in$ $ {\mathbb{R}^{12}}$, respectively. To obtain the matrix ${\vec{A}_m}$ in \eqref{eq:J4_11DOF_Am}, we can compute the terms in ${\vec{A}_m}$ as follows: let the $i$th cable-length vector be
\begin{align}
    {{{\vec L}}_i} = &{\;} {[a_{ix},a_{iy},a_{iz}]^T} - {[{p_{mx}},{p_{my}},{p_{mz}}]^T}- \vec{R}_g^m \nonumber\\
    &{\;}[r_{ix},r_{iy}, r_{iz}]^T,\; \{\forall \; i \in{\mathbb{N}} :1 \le i \le 12\},
    \label{eq:J4_appendix_hat_vec_Li}
\end{align}
with ${{{\vec L}}_i} \in {\mathbb{R}^3}$ denoting the position vector from the $i$th cable anchor point on the robot static frame to the $i$th cable anchor point on the mobile platform. Then, one can get the $i$th cable length
\begin{align}
    L_i  = &{\;} \lVert {[a_{ix},a_{iy},a_{iz}]^T}  -  {[{p_{mx}},{p_{my}},{p_{mz}}]^T}\nonumber\\
    &{\;} - \vec{R}_g^m{[r_{ix},r_{iy},r_{iz}]^T} \rVert,
    \label{eq:J4_appendix_Li}
\end{align}
 and the $i$th unit cable-length vector 
\begin{align}
    {{\hat {\vec{L}}}}_i=\frac{{\vec L}_i}{L_i} = {[{\hat L_{ix}},{\hat L_{iy}},{\hat L_{iz}}]^T} \in {\mathbb{R}^3}.
    \label{eq:J4_appendix_hat_Li}
\end{align}}

Furthermore, ${[r_{ix},r_{iy},r_{iz}]^T} \in {\mathbb{R}^3}$ denotes the position vector of the $i$th cable anchor point on the mobile platform with respect to the frame $\{O_m\}$, ${[a_{ix},a_{iy},a_{iz}]^T} \in {\mathbb{R}^3}$ represents the position vector of the $i$th cable anchor point on the robot static frame with respect to the frame $\{O\}$, and the parameters ${[r_{ix},r_{iy},r_{iz}]^T},{[a_{ix},a_{iy},a_{iz}]^T}$ are provided in~\cite{R.Qi2018j2}.

% ==================================================
% \section*{Acknowledgment}
% ==================================================
\section*{Acknowledgment}
The authors would like to knowledge the financial support of the Natural Sciences and Engineering Research Council of Canada (NSERC).

% \vspace{-0.6cm}
%==================================================
% References, \bibliography
% Note: I use "doi_disable=" to replace "doi=" if there is no "http:// ..."
%==================================================
\bibliographystyle{IEEEtran}
\bibliography{Bibliography,IEEEabrv}

% Generated by IEEEtran.bst, version: 1.14 (2015/08/26)
\begin{thebibliography}{10}
\providecommand{\url}[1]{#1}
\csname url@samestyle\endcsname
\providecommand{\newblock}{\relax}
\providecommand{\bibinfo}[2]{#2}
\providecommand{\BIBentrySTDinterwordspacing}{\spaceskip=0pt\relax}
\providecommand{\BIBentryALTinterwordstretchfactor}{4}
\providecommand{\BIBentryALTinterwordspacing}{\spaceskip=\fontdimen2\font plus
\BIBentryALTinterwordstretchfactor\fontdimen3\font minus
  \fontdimen4\font\relax}
\providecommand{\BIBforeignlanguage}[2]{{%
\expandafter\ifx\csname l@#1\endcsname\relax
\typeout{** WARNING: IEEEtran.bst: No hyphenation pattern has been}%
\typeout{** loaded for the language `#1'. Using the pattern for}%
\typeout{** the default language instead.}%
\else
\language=\csname l@#1\endcsname
\fi
#2}}
\providecommand{\BIBdecl}{\relax}
\BIBdecl

\bibitem{Mendez2014}
S.~J.~T. {M{\'e}ndez}, ``{Low Mobility Cable Robot with Application to Robotic
  Warehousing},'' Ph.D. dissertation, University of Waterloo, Waterloo, ON,
  Canada, 2014.

\bibitem{H.Jamshidifar2017}
H.~Jamshidifar, A.~Khajepour \emph{et~al.}, ``Kinematically-constrained
  redundant cable-driven parallel robots: Modeling, redundancy analysis, and
  stiffness optimization,'' \emph{IEEE/ASME Trans. Mechatronics}, vol.~22,
  no.~2, pp. 921--930, April 2017.

\bibitem{H.Jamshidifar2018}
H.~Jamshidifar, S.~Khosravani, B.~Fidan, and A.~Khajepour, ``Vibration
  decoupled modeling and robust control of redundant cable-driven parallel
  robots,'' \emph{IEEE/ASME Trans. Mechatronics}, vol.~23, no.~2, pp. 690--701,
  April 2018.

\bibitem{Z.Mu2018}
Z.~{Mu}, H.~{Yuan} \emph{et~al.}, ``{A Segmented Geometry Method for Kinematics
  and Configuration Planning of Spatial Hyper-Redundant Manipulators},''
  \emph{{IEEE} Trans. Syst., Man, Cybern., Syst.}, pp. 1--11, 2018.

\bibitem{M.Chen2018}
M.~{Chen}, Y.~{Ren}, and J.~{Liu}, ``{Antidisturbance Control for a Suspension
  Cable System of Helicopter Subject to Input Nonlinearities},'' \emph{{IEEE}
  Trans. Syst., Man, Cybern., Syst.}, vol.~48, no.~12, pp. 2292--2304, Dec.
  2018.

\bibitem{Otis2009}
M.~J.~D. {Otis}, S.~{Perreault} \emph{et~al.}, ``{Determination and Management
  of Cable Interferences Between Two 6-DOF Foot Platforms in a Cable-Driven
  Locomotion Interface},'' \emph{{IEEE} Trans. Syst., Man, Cybern. {A}},
  vol.~39, no.~3, pp. 528--544, May 2009.

\bibitem{Chiaverini97}
S.~{Chiaverini}, ``Singularity-robust task-priority redundancy resolution for
  real-time kinematic control of robot manipulators,'' \emph{IEEE Trans. Robot.
  Autom.}, vol.~13, no.~3, pp. 398--410, June 1997.

\bibitem{R.Tedrake2019}
\BIBentryALTinterwordspacing
R.~{Tedrake}, ``{Underactuated Robotics: Algorithms for Walking, Running,
  Swimming, Flying, and Manipulation (Course Notes for MIT 6.832)},''
  [Accessed: March 11, 2020]. [Online]. Available:
  \url{http://underactuated.mit.edu/}
\BIBentrySTDinterwordspacing

\bibitem{ARSENAULT20131}
M.~Arsenault, ``Workspace and stiffness analysis of a three-degree-of-freedom
  spatial cable-suspended parallel mechanism while considering cable mass,''
  \emph{Mechan. Mach. Theory}, vol.~66, pp. 1--13, 2013.

\bibitem{LIM20111265}
W.~B. {Lim}, G.~{Yang} \emph{et~al.}, ``A generic force-closure analysis
  algorithm for cable-driven parallel manipulators,'' \emph{Mechan. Mach.
  Theory}, vol.~46, no.~9, pp. 1265--1275, 2011.

\bibitem{J.Li2017}
J.~Li, S.~Andrews \emph{et~al.}, ``{Task-based Design of Cable-driven
  Articulated Mechanisms},'' in \emph{Proc. 1st Annual ACM Sym. Comp. Fabri.},
  New York, NY, USA, 2017, pp. 6:1--6:12.

\bibitem{Wolovich1984}
W.~A. {Wolovich} and H.~{Elliott}, ``A computational technique for inverse
  kinematics,'' in \emph{Proc. 23rd IEEE Conf. Decision Control}, Dec. 1984,
  pp. 1359--1363.

\bibitem{Wang1991}
L.~T. {Wang} and C.~C. {Chen}, ``A combined optimization method for solving the
  inverse kinematics problems of mechanical manipulators,'' \emph{IEEE Trans.
  Robot. Autom.}, vol.~7, no.~4, pp. 489--499, Aug. 1991.

\bibitem{Wampler1986}
C.~W. {Wampler}, ``Manipulator inverse kinematic solutions based on vector
  formulations and damped least-squares methods,'' \emph{{IEEE} Trans. Syst.,
  Man, Cybern.}, vol.~16, no.~1, pp. 93--101, Jan. 1986.

\bibitem{Nakamura1986}
Y.~{Nakamura} and H.~{Hanafusa}, ``Inverse kinematic solutions with singularity
  robustness for robot manipulator control,'' \emph{ASME. J. Dyn. Sys., Meas.,
  Control}, vol. 108, no.~3, pp. 163--171, Sep. 1986.

\bibitem{Zhao1994}
J.~{Zhao} and N.~I. {Badler}, ``Inverse kinematics positioning using nonlinear
  programming for highly articulated figures,'' \emph{ACM Trans. Graph.},
  vol.~13, no.~4, pp. 313--336, Oct. 1994.

\bibitem{Barrette2005}
G.~{Barrette} and C.~M. {Gosselin}, ``Determination of the dynamic workspace of
  cable-driven planar parallel mechanisms,'' \emph{ASME. J. Mech. Des.}, vol.
  127, no.~2, pp. 242--248, March 2005.

\bibitem{Bruckmann2006}
T.~Bruckmann, A.~Pott, and M.~Hiller, ``Calculating force distributions for
  redundantly actuated tendon-based stewart platforms,'' in \emph{Advances in
  Robot Kinematics}, J.~Lennar{\v{c}}i{\v{c}} and B.~Roth, Eds.\hskip 1em plus
  0.5em minus 0.4em\relax Dordrecht: Springer Netherlands, 2006, pp. 403--412.

\bibitem{Y.Zhang2007}
Y.~{Zhang} and S.~{Ma}, ``Minimum-energy redundancy resolution of robot
  manipulators unified by quadratic programming and its online solution,'' in
  \emph{Proc. Int. Conf. Mechatronics Autom.}, Harbin, China, Aug. 2007, pp.
  3232--3237.

\bibitem{Flacco2015}
F.~{Flacco} and A.~D. {Luca}, ``Discrete-time redundancy resolution at the
  velocity level with acceleration/torque optimization properties,''
  \emph{Robot. Auton. Syst.}, vol.~70, pp. 191--201, 2015.

\bibitem{Tang2001}
W.~S. {Tang} and J.~{Wang}, ``A recurrent neural network for minimum
  infinity-norm kinematic control of redundant manipulators with an improved
  problem formulation and reduced architecture complexity,'' \emph{{IEEE}
  Trans. Syst., Man, Cybern. {B}}, vol.~31, no.~1, pp. 98--105, Feb. 2001.

\bibitem{Zhang2006}
Y.~Zhang, ``A set of nonlinear equations and inequalities arising in robotics
  and its online solution via a primal neural network,'' \emph{Neurocomputing},
  vol.~70, no.~1, pp. 513--524, 2006.

\bibitem{Y.Zhang2004}
Y.~{Zhang}, S.~S. {Ge}, and T.~H. {Lee}, ``A unified
  quadratic-programming-based dynamical system approach to joint torque
  optimization of physically constrained redundant manipulators,'' \emph{{IEEE}
  Trans. Syst., Man, Cybern. {B}}, vol.~34, no.~5, pp. 2126--2132, Oct. 2004.

\bibitem{Z.Zhang2018}
Z.~{Zhang}, Y.~{Lin} \emph{et~al.}, ``Tricriteria optimization-coordination
  motion of dual-redundant-robot manipulators for complex path planning,''
  \emph{IEEE Trans. Control Syst. Technol.}, vol.~26, no.~4, pp. 1345--1357,
  July 2018.

\bibitem{Y.Zhang2006}
Y.~Zhang, ``Inverse-free computation for infinity-norm torque minimization of
  robot manipulators,'' \emph{Mechatronics}, vol.~16, no.~3, pp. 177--184,
  2006.

\bibitem{Guo2012}
D.~{Guo} and Y.~{Zhang}, ``Different-level two-norm and infinity-norm
  minimization to remedy joint-torque instability/divergence for redundant
  robot manipulators,'' \emph{Robot. Auton. Syst.}, vol.~60, no.~6, pp.
  874--888, 2012.

\bibitem{Tang1999}
W.~S. {Tang}, J.~{Wang}, and Y.~{Xu}, ``Infinity-norm torque minimization for
  redundant manipulators using a recurrent neural network,'' in \emph{Proc.
  38th IEEE Conf. Decision Control}, vol.~3, Phoenix, Arizona, USA, Dec. 1999,
  pp. 2168--2173.

\bibitem{Kazerounian1988}
K.~{Kazerounian} and Z.~{Wang}, ``Global versus local optimization in
  redundancy resolution of robotic manipulators,'' \emph{Int. J. Robot.
  Research}, vol.~7, no.~5, pp. 3--12, 1988.

\bibitem{S.Ma1994}
S.~Ma, ``Local torque optimization of redundant manipulators in torque-based
  formulation,'' in \emph{Proc. 20th Conf. IEEE Ind. Electron.}, vol.~2,
  Bologna, Italy, Sep. 1994, pp. 697--702.

\bibitem{Rushton2016}
M.~Rushton, ``Vibration control in cable robots using a multi-axis reaction
  system,'' Master's thesis, University of Waterloo, Waterloo, ON, Canada,
  2016.

\bibitem{Rushton2018}
M.~Rushton and A.~Khajepour, ``Transverse vibration control in planar
  cable-driven robotic manipulators,'' in \emph{Cable-Driven Parallel Robots},
  C.~Gosselin, P.~Cardou, T.~Bruckmann, and A.~Pott, Eds.\hskip 1em plus 0.5em
  minus 0.4em\relax Cham: Springer International Publishing, 2018, pp.
  243--253.

\bibitem{R.Qi2018j2}
R.~{Qi}, M.~{Rushton}, A.~{Khajepour}, and W.~W. {Melek}, ``{Decoupled modeling
  and model predictive control of a hybrid cable-driven robot (HCDR)},''
  \emph{Robot. Auton. Syst.}, vol. 118, pp. 1--12, 2019.

\bibitem{R.Qi2019j3}
R.~Qi, A.~Khajepour, and W.~W. Melek, ``Modeling, tracking, vibration and
  balance control of an underactuated mobile manipulator {(UMM)},''
  \emph{Control Eng. Pract.}, vol.~93, p. 104159, 2019.

\bibitem{R.Qi2019j4}
R.~{Qi}, A.~{Khajepour}, and W.~W. {Melek}, ``{Generalized Flexible Hybrid
  Cable-Driven Robot (HCDR): Modeling, Control, and Analysis},'' 2019,
  arXiv:1911.06222.

\bibitem{Luenberger2008}
D.~G. Luenberger and Y.~Ye, \emph{Linear Nonlinear Programming}, 3rd~ed.\hskip
  1em plus 0.5em minus 0.4em\relax New York, NY, USA: Springer, 2008.

\bibitem{W.He2020}
W.~{He}, C.~{Xue}, X.~{Yu}, Z.~{Li}, and C.~{Yang}, ``Admittance-based
  controller design for physical human-robot interaction in the constrained
  task space,'' \emph{{IEEE} Trans. Autom. Sci. Eng.}, pp. 1--13, 2020.

\bibitem{L.Kong2019}
L.~{Kong}, W.~{He}, C.~{Yang}, Z.~{Li}, and C.~{Sun}, ``Adaptive fuzzy control
  for coordinated multiple robots with constraint using impedance learning,''
  \emph{{IEEE} Trans. Cybern.}, vol.~49, no.~8, pp. 3052--3063, 2019.

\bibitem{L.Kong20192}
L.~{Kong}, W.~{He}, Y.~{Dong}, L.~{Cheng}, C.~{Yang}, and Z.~{Li}, ``Asymmetric
  bounded neural control for an uncertain robot by state feedback and output
  feedback,'' \emph{{IEEE} Trans. Syst., Man, Cybern., Syst.}, pp. 1--12, 2019.

\bibitem{Nabat2005}
V.~{Nabat}, M.~{de la O Rodriguez} \emph{et~al.}, ``Par4: very high speed
  parallel robot for pick-and-place,'' in \emph{Proc. IEEE/RSJ Int. Conf.
  Intel. Robot. Syst.}, Edmonton, Alta., Canada, Aug. 2005, pp. 553--558.

\end{thebibliography}

\vfill
\end{document}